\documentclass[10pt,twocolumn,letterpaper]{article}

\usepackage{cvpr}              % To produce the CAMERA-READY version
\usepackage{graphicx}
\usepackage{amsmath}
\usepackage{amssymb}
\usepackage{booktabs}

\usepackage{times}
\usepackage{epsfig}
\usepackage{color}
\usepackage{multirow}
\usepackage{enumitem}
\usepackage[labelformat=simple]{subcaption}

\usepackage{array}
\newcommand{\PreserveBackslash}[1]{\let\temp=\\#1\let\\=\temp}
\newcolumntype{C}[1]{>{\PreserveBackslash\centering}p{#1}}
\usepackage{makecell}

\usepackage[pagebackref,breaklinks,colorlinks]{hyperref}

\usepackage[capitalize]{cleveref}
\crefname{section}{Sec.}{Secs.}
\Crefname{section}{Section}{Sections}
\Crefname{table}{Table}{Tables}
\crefname{table}{Tab.}{Tabs.}

\def\cvprPaperID{9} % *** Enter the CVPR Paper ID here
\def\confName{CVPR Workshop - StruCo3D}
\def\confYear{2023}

\begin{document}

%%%%%%%%% TITLE - PLEASE UPDATE
\title{\vspace{-5pt} Attention-based Part Assembly for 3D Volumetric Shape Modeling}

\author{Chengzhi Wu$^{1}$ \quad Junwei Zheng$^{1}$
\quad Julius Pfrommer$^{2,3}$ \quad Jürgen Beyerer$^{1,2,3}$ \vspace{1pt}
\and
$^{1}$Karlsruhe Institute of Technology, Germany \quad
$^{2}$Fraunhofer IOSB, Germany\\
$^{3}$Fraunhofer Center for Machine Learning, Germany \vspace{-1pt}
\and
{\tt\footnotesize
%\tt\fontsize{7.5pt}{6pt}\selectfont
\{chengzhi.wu, junwei.zheng\}@kit.edu, \quad \{julius.pfrommer, juergen.beyerer\}@iosb.fraunhofer.de}
}

\maketitle

%%%%%%%%% ABSTRACT
\begin{abstract}
Modeling a 3D volumetric shape as an assembly of decomposed shape parts is much more challenging, but semantically more valuable than direct reconstruction from a full shape representation. The neural network needs to implicitly learn part relations coherently, which is typically performed by dedicated network layers that can generate transformation matrices for each part. In this paper, we propose a VoxAttention network architecture for attention-based part assembly. We further propose a variant of using channel-wise part attention and show the advantages of this approach. Experimental results show that our method outperforms most state-of-the-art methods for the part relation-aware 3D shape modeling task.
\end{abstract}

%%%%%%%%% BODY TEXT
\section{Introduction}
\label{sec:introduction}
Modeling of 3D shapes has been a central research topic in the computer graphics domain for decades. However, although great progress has been made since the seminal work of ShapeNet \cite{Chang2015ShapeNetAI}, most existing 3D modeling methods only focus on full shape modeling and are part relation-oblivious \cite{Brock2016GenerativeAD,Wu2016LearningAP,Girdhar2016LearningAP}. Since no semantic information is used in those methods, details of the 3D shapes, especially those of small volume yet exquisite parts, are often badly generated in a blurred or shattered manner.

Recently, part-based 3D shape modeling is attracting increasing research interest since it generates better part details and gives more insight into part relations. Those methods usually consist of two steps, shape parts generation and part assembly. Generating shape parts, similar to generating a full shape, is easy to achieve with 3D generative neural networks (either AE, VAE, GAN, or their combinations). 
However, how to encode and learn part relations for assembling the full shape is still an open question. Current state-of-the-art methods either encode all parts together and then decode it to a full shape \cite{Wang2018GlobaltolocalGM,Wu2019StructureawareGN}, or use simple regression layers to get part transformation matrices \cite{Li2020LearningPG}. However, they only focus on how to integrate all part feature information during the learning, but neglect to maintain their independence at the same time.

In this paper, we propose a part-based attention network to learn part relations from their features in different network layers, including the vector representation, feature maps in the generator, or even the final decoded output. The part dimension will be preserved throughout the training. The attention mechanism, which originated from Transformer \cite{Vaswani2017AttentionIA}, now is widely used in the computer vision domain and has proven its effectiveness in various frameworks including iGPT \cite{Chen2020GenerativePF}, ViT \cite{Dosovitskiy2021AnII}, IPT \cite{Chen2021PreTrainedIP}, etc. However, it is mostly applied to 2D data, \ie, images currently. For 3D data, most attention-based works are on the point cloud data representation \cite{Guo2021PCTPC,Zhao2021PointT,Engel2021PointT}, 3D volumetric data have been rarely explored.
Like that there are several different levels of attention on 2D data which include pixel-based (smallest input entity), patch-based (enlarged the perception field), and part-based (with semantic information), for 3D volumetric data, we can also apply the attention to levels of voxel-based, patch-based, or part-based. In our case, since we are interested in the part relations, we use part-based attention for the 3D shape modeling task. 

The proposed VoxAttention framework first learns an autoencoder to encode the full shape into coherent yet mutual orthogonal part latent representations, and decode them into semantic parts. Secondly, transformation matrices for those decoded parts are learned by applying attention to their feature information in different layers to assemble them. Our experimental results demonstrate that the proposed method achieves excellent performance both qualitatively and quantitatively on the 3D shape modeling task.

Our main contributions include:
\begin{itemize}[itemsep=-3pt,topsep=-5pt,left=3pt]
\item A part-based attention neural network to learn semantic part relations for better 3d shape assembly. 
\item An optional channel-wise attention strategy on top of the normal part attention model for feature learning.
\item An additional attention consistency loss to prevent the network from mode collapse when multiple feature layers are used for computing the part relations.
\end{itemize}

\section{Related Work}
\label{sec:relatedWork}
\subsection{Generative Networks for 3D Shape Modeling}
With the development of deep learning techniques, there has been a surge of research interest in deep generative models in recent years. Since the seminal work of ShapeNet \cite{Chang2015ShapeNetAI}, which is a large 3D shape model dataset that is publicly available, lots of research has been conducted in building 3D generative models for modeling 3D shapes. Brock \etal \cite{Brock2016GenerativeAD} use a 3D VAE neural network to encode and decode 3D shapes. Wu \etal use a 3D-GAN \cite{Wu2016LearningAP} to generate 3D shapes from latent vectors. Additional image information has also been used in various methods. TL-network \cite{Girdhar2016LearningAP} first applies an autoencoder on 3D shapes, then forces the image encoder to learn a similar latent representation in between, while SwitchVAE \cite{Wu2022SwitchVAE} uses a switch scheme to allow both branches updates coherently. Wu \etal \cite{Wu20153DSA} decrease the uncertainty of 3D voxel grids by predicting the next best viewing angle. View images have also been used to compute an additional loss in \cite{Tulsiani2018MultiviewCA}. 3D-R2N2 \cite{Choy20163DR2N2AU} uses a recurrent neural network to encode multi-view information for 3D shape modeling. Octree-based methods \cite{wang2017cnn,Riegler2017OctNetLD} have been proposed for more computationally efficient modeling. Apart from the volumetric data representation, 3D generative modeling methods have also been proposed on other data representations, \eg, point clouds \cite{Achlioptas2018LearningRA,Yang2019PointFlow3P,Pumarola2020CFlowCG,Valsesia2019LearningLG}, surface meshes \cite{Wang2018Pixel2MeshG3,Wen2019Pixel2MeshM3,Henderson2020Leveraging2D}, or even implicit fields \cite{Chen2019LearningIF,Park2019DeepSDFLC,Mescheder2019OccupancyNL}.

\subsection{Assembly-based 3D Shape Modeling}
Compared to generating a full 3D shape directly, generating shape semantic parts separately and assembling them is more valuable since it considers the structural decomposition of 3D shapes and the connections between shape parts. Li \etal \cite{Li2017GRASSGR} propose the first deep structure-aware modeling framework GRASS for 3D shapes by employing a recursive neural network. StructureNet \cite{Mo2019StructureNetHG} also uses a tree-based architecture to generate shape parts in sequences by extending the binary tree to N-ary tree. 3D-PRNN \cite{Zou20173DPRNNGS} and SCORES \cite{Zhu2018SCORESSC} use a similar method but only shape primitives are generated. StructureEdit \cite{Mo2020StructEditLS} further provides a framework for easy editing. With additional structure landmarks, Balashova  \cite{Balashova2018StructureAwareSS} adds a structure detector in combination with the generator for better shape modeling. 

Apart from the above methods that are mostly structure tree-based, there are lots of methods that learn part relations directly. COALESCE \cite{Yin2020COALESCECA} assembles shape components by learning to synthesize connections. G2LGAN \cite{Wang2018GlobaltolocalGM} sends all the generated parts into another autoencoder to decode out a full shape. SAGNet \cite{Wu2019StructureawareGN} uses two branches for encoding the part and structure information separately.
For methods that learn spatial transformations for the parts directly, CompoNet \cite{Schor2019CompoNetLT} learns the transformations from latent representations of 3D point clouds, but its output shapes are often disconnected. PQ-Net \cite{Wu2020PQNETAG} employs a Seq2Seq generative network to reconstruct 3D shapes part by part with gated recurrent units. Dubrovina \etal \cite{Dubrovina2019CompositeSM} propose a decomposer-composer network to learn a factorized embedding space. 
A more recent work of PAGENet \cite{Li2020LearningPG} learns part transformations only from decoded outputs with a special augmented dataset. All the above methods merge the part feature information together at a certain step. We instead propose to keep the part dimension intact and use an attention-based method to learn part relations.

\begin{figure}[t]
\centering
\includegraphics[width=\linewidth,trim=2 2 2 2,clip]{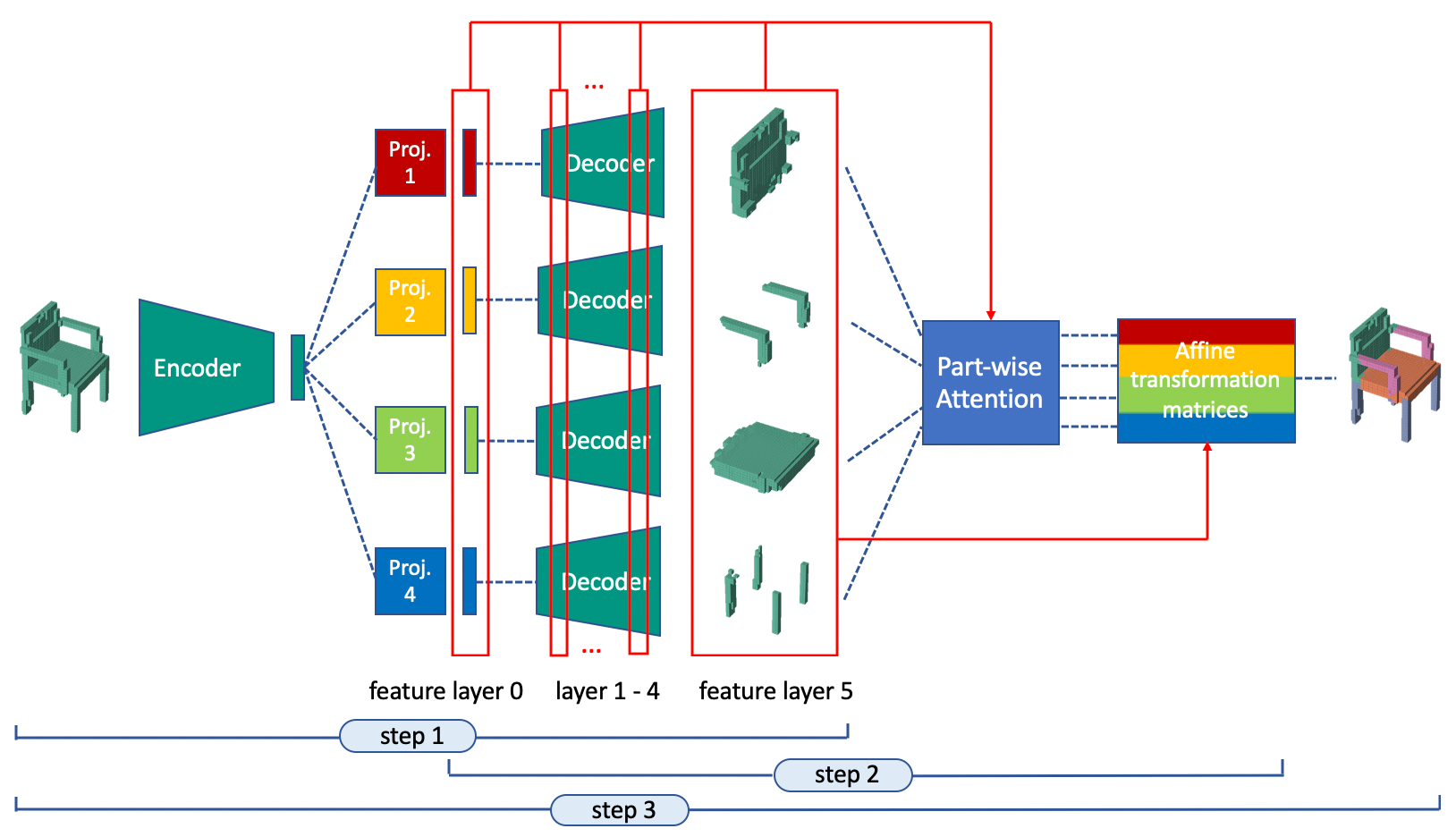}
\caption{Pipeline of proposed VoxAttention framework. Single or multiple feature layers from the decoder are used to learn relative relations for shape parts with part-based attention.  \vspace{-0.cm}}
\label{fig:pipeline}
\end{figure}

\subsection{Attention in Computer Vision}
Attention-based methods have almost dominated the natural language processing (NLP) domain after the far-reaching model Transformer \cite{Vaswani2017AttentionIA} was proposed. In recent years, researchers have applied the attention mechanism to computer vision tasks and achieved great success. DERT \cite{Carion2020EndtoEndOD} applies attention to flattened feature maps that are convoluted from a CNN for object detection. To achieve better self-supervised learning on images, iGPT \cite{Chen2020GenerativePF} masks some pixels out and uses an attention-based network for generative pre-training. ViT \cite{Dosovitskiy2021AnII} cuts an input image into patches and learns over those flattened patches with a transformer encoder for classification. IPT \cite{Chen2021PreTrainedIP} does pre-training on multiple tasks, also with a transformer encoder. A good survey on visual transformers is given in \cite{Han2020ASO}. However, most current visual transformers only apply attention to the 2D image data. For 3D data, the researches that have adopted the attention mechanism are  mostly on point clouds \cite{Guo2021PCTPC, Zhao2021PointT, Engel2021PointT}.
There are also works that voxelize the whole point cloud space to use an attention-based method on the voxel level for better 3D detection \cite{Mao2021VoxelTF,He2020SVGANetSV}. However, to our knowledge, there are few works that apply attention directly to shapes of 3D volumetric data, especially with regard to the semantic part dimension.

\section{Methodology}
\label{sec:methodology}
\subsection{Overall Framework}
Methods dealing with the part assembly task mostly follow a same routine which consists of two steps of generating shape parts and learning part relations \cite{Wang2018GlobaltolocalGM, Li2020LearningPG}. 
Our VoxAttention framework also follows this pattern but with an additional fine-tuning step added afterward as illustrated in Figure \ref{fig:pipeline}. An encoder-decoder architecture is used first to learn part generation. Then the latent representation, feature maps in each decoder layer, and decoded output of each part are used for part-based self-attention to learn part relative relations. After applying the learned affine transformation matrices, a full 3D shape is finally assembled. For each step, different losses are used for the training while different metrics are used to evaluate the learning performance.

\subsection{Part Generator}
The first step trains an autoencoder for 3D shape parts reconstruction. Although we want to obtain feature information in a part-wise manner, the information should still keep related implicitly in the latent space. Using separate autoencoders for each part as in \cite{Li2020LearningPG} will not work since they have no information communication. The part feature information needs to be learned jointly. Here, to generate parts separately yet coherently, a shared autoencoder is used to reconstruct each part.  We adopt the method in \cite{Dubrovina2019CompositeSM} to achieve the decomposition of shape latent representations. Mutual orthogonal projection matrices are learned for each part respectively.
Denote $N_p$ as the number of parts a shape category contains (\eg, a chair contains four parts), the projection matrices $P_i (i=1,\dots,N_p)$ satisfy the properties of (i) projection property: $P_i^2 = P_i$, (ii) mutual orthogonal: $P_iP_j = \textit{0} \ (i \neq j)$, and (iii) forming an identity matrix together: $\sum_{i=1}^{N_p} P_i = I$. 
A partition of the identity (PI) loss $L_\text{PI}$ is used to measure the deviation of the learned projection matrices $P_1, \dots, P_{N_p}$ from the optimal projection:
\begin{equation}
L_\text{PI} = \sum_{i=1}^{N_p}||P_i^2 - P_i||_F^2 
+ \sum_{i,j=1,i\neq j}^{N_p}||P_i P_j||_F^2 
+ ||\sum_{i=1}^{N_p}P_i - I||_F^2
\end{equation}
For the part reconstruction loss $L_\text{part}$, same as \cite{Brock2016GenerativeAD,Wu2022SwitchVAE}, a modified binary cross entropy loss is used by introducing a hyper-parameter $\gamma$, which weights the relative importance of false positives against false negatives.
\begin{equation}
L_\text{part} = -2(\gamma t\log(o) + (1-\gamma)(1-t)\log(1-o))
\end{equation}
where $t$ is the target value in $\{0, 1\}$ and $o$ is the output of the network in $(0,1)$ at each output element.
Adding weights to different loss terms, the total loss in step 1 is defined as:
\begin{equation}
L_\text{s1} = \omega_\text{PI} L_\text{PI} + \omega_\text{part} L_\text{part}
\end{equation}
For evaluation, part mIoU is used as the metric for step 1.

\begin{figure}[t]
\centering
\includegraphics[width=\linewidth,trim=2 2 2 2,clip]{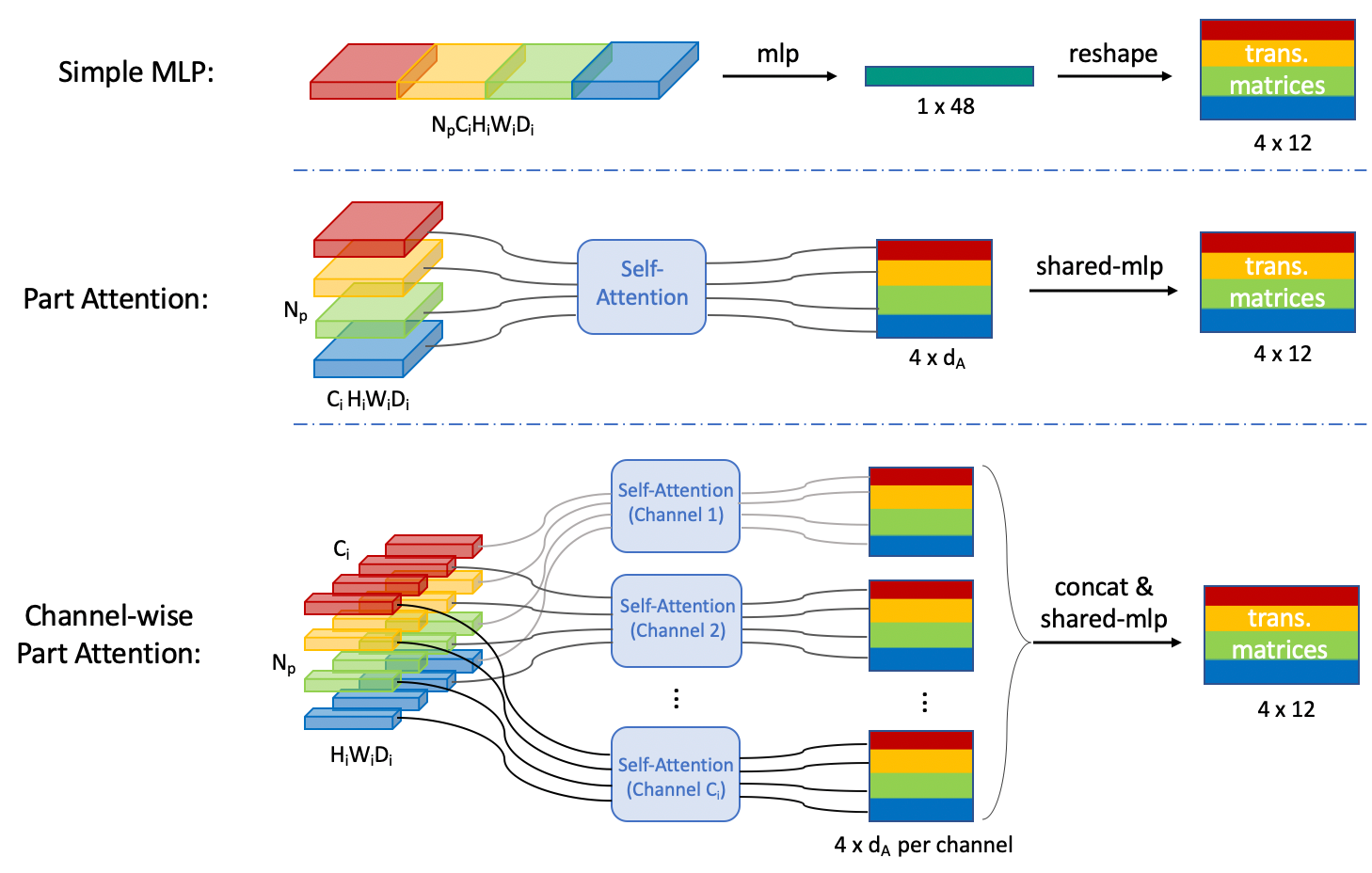}
\caption{An illustrative comparison figure of three different methods, using an example of $N_p =4$ and only latent features from a single $i$th feature layer is used as input.  \vspace{-0.cm}}
\label{fig:3methods}
\end{figure}

\begin{figure}[t]
\centering
%\vspace{-0.2cm}
\includegraphics[width=\linewidth,trim=2 2 2 2,clip]{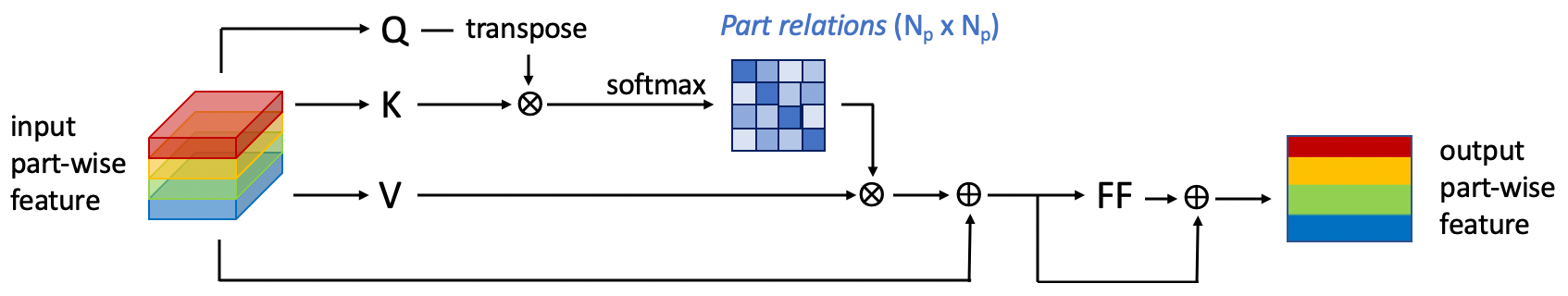}
\caption{One attention block used for learning part relations. Q, K, and V mean query, key, and value respectively. FF means feed-forward network.   \vspace{-0.cm}}
\label{fig:attenBlock}
\end{figure}

\begin{figure*}[t]
\centering
\includegraphics[width=0.95\linewidth,trim=2 8 2 2, clip]{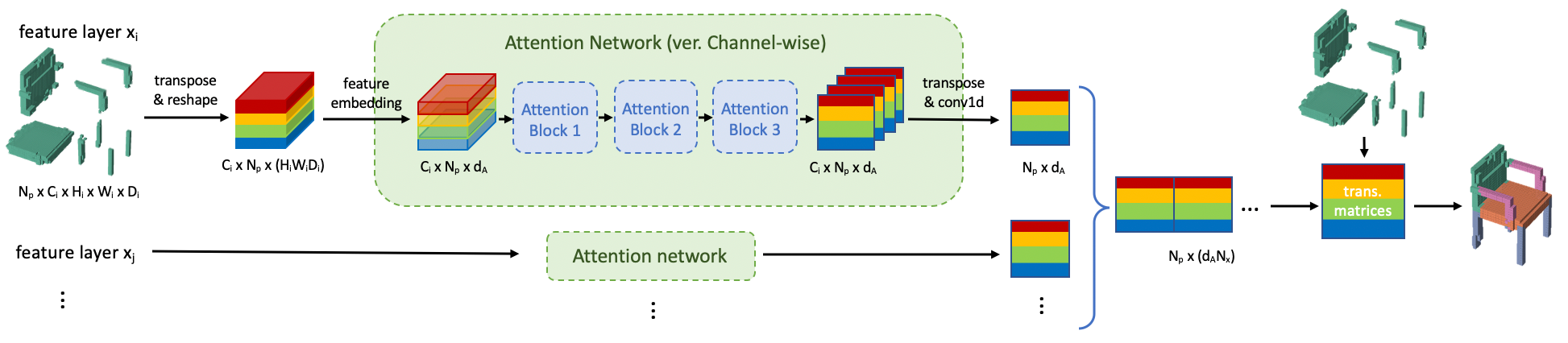}
\caption{A detailed overview of proposed attention network architecture (channel-wise part attention version). Multiple feature layers are used as input. $N_x$ stands for the total number of input feature layers. \vspace{-0.1cm}}
\label{fig:attenNet}
\end{figure*}

\subsection{Learning Transformation Matrices}
\label{sec:attention}
The second step is the key for part assembly. A simple idea is to apply multi-layer perceptron (MLP) directly on the decoded parts and latent representations. 
We instead propose a part attention-based method for learning better part relations.
Figure \ref{fig:3methods} illustrates how three methods process the input features differently. 
$d_A$ stands for the embedding dimension for attention blocks if they are used.

Assuming the shape has $N_p$ parts, in the $i$th feature layer of the decoder, it has $C_i$ feature channels and the feature maps are of a resolution of $H_i \times W_i \times D_i$. When this single feature layer is used in this step, ignoring the batch size dimension, the input feature is a 5-dimensional tensor of $N_p \times C_i \times H_i \times W_i \times D_i$. 
\textbf{Simple MLP} method reshapes the input feature into a 1-dimensional vector of $(N_p C_i H_i W_i D_i)$ and applies MLP on it directly. The output vector is then reshaped to the shape of transformation matrices. \textbf{Part Attention} method reshapes the input feature into a 2-dimensional tensor of $N_p \times (C_i H_i W_i D_i)$ and applies self-attention along the part dimension. An attention block is illustrated in Figure \ref{fig:attenBlock}. With part relations learned during the attention operation, parts information is cross-communicated thus more comprehensive part-wise features are learned.  
A shared MLP is subsequently applied on all parts to learn their respective transformation matrices. \textbf{Channel-wise Part Attention} method further preserves the feature channel dimension and reshapes the input into a 3-dimensional tensor of $N_p \times C_i \times (H_i W_i D_i)$. For each channel, an attention block is applied to learn separate part features. All output features are subsequently concatenated and a shared MLP is applied to learn part transformation matrices.

In the actual application, since the original full feature dimension is too large to use directly, in order to reduce the feature dimension and keep it consistent for all input feature layers, we first embed the original features to the attention embedding dimension $d_A$ before performing the attention operation. To be more specific, the input feature is embedded into $N_p \times d_A$ for the normal part attention case, while into $C_i \times N_p \times d_A$ for the channel-wise part attention case.
After the embedding, three consecutive attention blocks are applied in our actual framework.
Additionally, multiple feature layers can be used as input.
Figure \ref{fig:attenNet} gives an overview of the whole pipeline for learning transformation matrices, using an example of the channel-wise part attention version and with multiple feature layers as input.
Each feature layer is firstly reshaped and embedded into a fixed-dimension vector part-and-channel-wise, followed by three consecutive attention blocks. The outputs are then concatenated and a shared MLP is applied to learn respective part transformation matrices.
See more details of the network architectures in subsection \ref{sec:network_details}.

Note that the channel-wise strategy is not applicable to all attention-based modules. Take our task as an example, if different part features are decoded from non-shared decoders, \ie, multiple decoders that have a same structure but are separately trained and not parameter-shared as in \cite{Li2020LearningPG}, the channel-wise strategy is not feasible since the feature channels of a same channel index from different part decoders are irrelevant. In our case, a parameter-shared decoder is used for all different parts. This makes the channel-wise part attention feasible since part feature information is still implicitly related to each other on each feature channel in every feature layer.

%------------ detailed network architecture ---------
\begin{figure}[t]
\centering
\includegraphics[width=\linewidth,trim=2 2 2 2,clip]{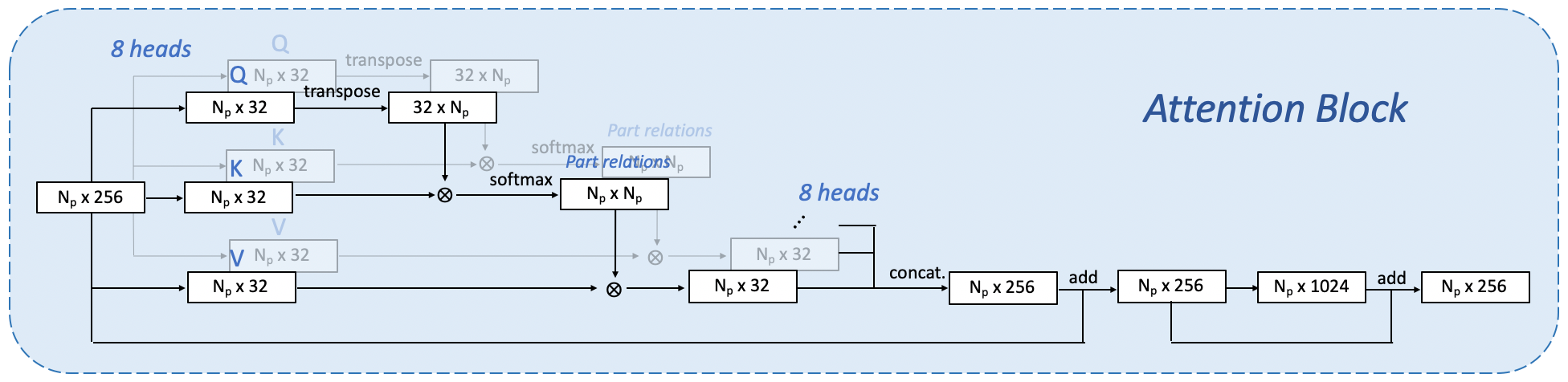}
\caption{Detailed network design of an attention block.\vspace{-0.1cm}}
\label{fig:block}
\end{figure}

\begin{figure}[t]
\centering
\includegraphics[width=\linewidth,trim=2 2 2 2,clip]{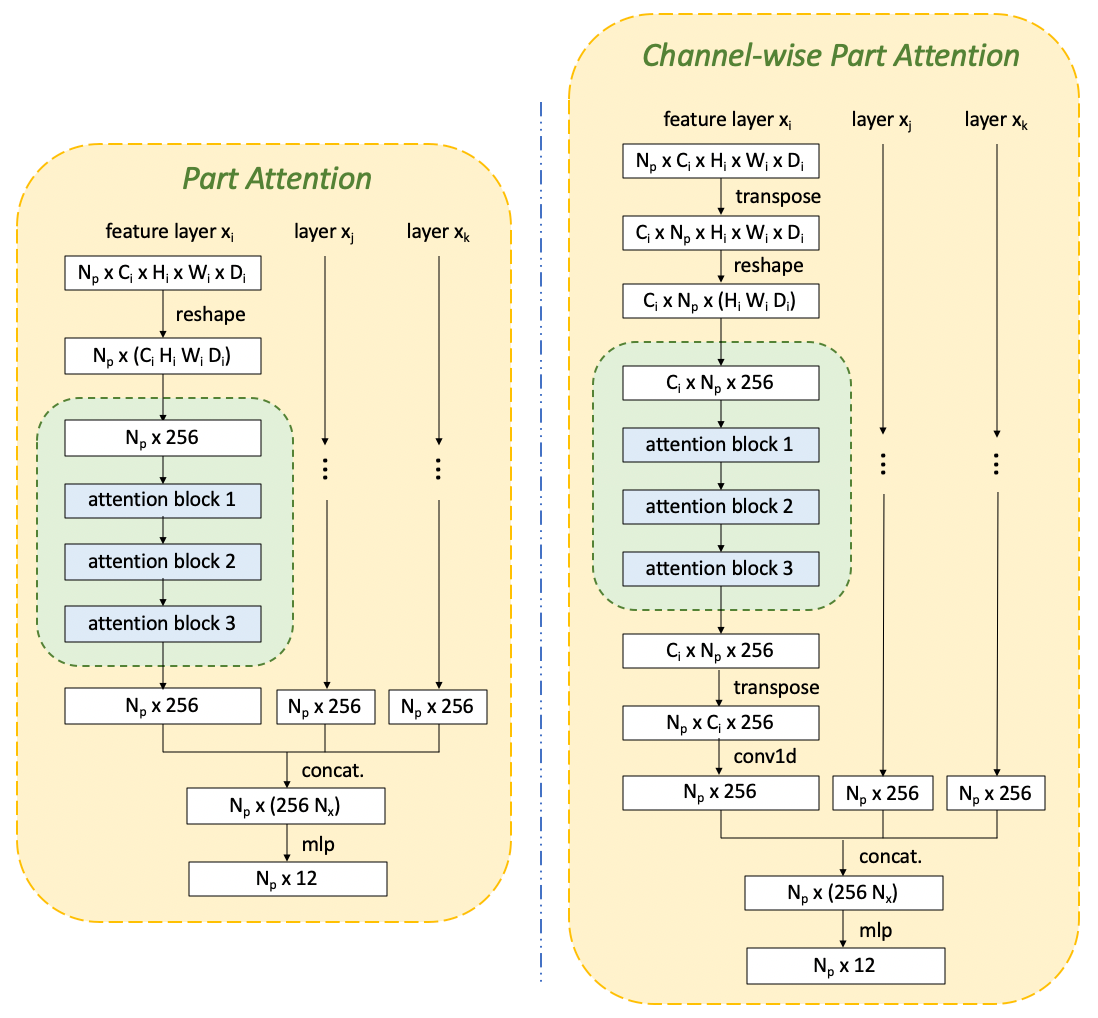}
\caption{Detailed network design of proposed network architectures. Both part attention and channel-wise part attention models are presented. $N_p$ stands for the number of parts of the input category. $N_x$ stands for the total number of input feature layers. \vspace{-0.2cm}}
\label{fig:net}
\end{figure}
%--------- detailed network architecture (end) ---------

In step 2, the transformation loss $L_\text{trans}$ is an $L2$ loss defined between the predicted and the ground truth transformation matrix parameters, summed over all $N_p$ parts. Additionally, if multiple feature layers are used as input for the attention network, an optional attention consistency loss $L_\text{AC}$ may be added. The attention consistency loss is defined as the $L2$ difference between the outputs of the attention network of all feature layers before concatenating them.
Adding weights to different loss terms, the total loss in step 2 is defined as:
\begin{equation}
L_\text{s2} = \omega_\text{trans} L_\text{trans} + \omega_\text{AC} L_\text{AC}
\end{equation}
For evaluation, transformation matrix MSE and full shape mIoU are used as the evaluation metrics for step 2. 

\subsection{Fine-tuning}
Our network is a complex one and end-to-end training does not work well. Thus we divide our training into several steps like many other related works \cite{Wang2018GlobaltolocalGM, Li2020LearningPG}. In step 1, only weights in the encoder-decoder are trained. In step 2, the weights in encoder-decoder are frozen and only weights in the part-based attention network are trained. In step 3, all weights are fine-tuned with additional training.

The fine-tuning step uses all the losses above, with an additional full shape reconstruction $L_\text{shape}$ loss added. $L_\text{shape}$ uses a same formula as $L_\text{part}$ but computes over the whole shape.
To achieve better fine-tuning, weights are used for all terms. The total loss in step 3 is defined as:
\begin{equation}
L_\text{s3} = \omega_\text{PI} L_\text{PI} + \omega_\text{part} L_\text{part} + \omega_\text{trans} L_\text{trans} + \omega_\text{AC} L_\text{AC} + \omega_\text{shape} L_\text{shape}
\end{equation}
For evaluation, all the above evaluation metrics, part mIoU, transformation matrices MSE, and shape mIoU are used.

\section{Experiments}
\label{sec:experiments}
\subsection{Dataset and Network Details}
\label{sec:network_details}
In our experiments, we use the ShapeNet Part dataset \cite{Yi2016ASA}, which is a subset of the ShapeNet dataset \cite{Chang2015ShapeNetAI} with part annotations. Shapes are firstly converted into voxel grids of resolution $32^3$ using binvox \cite{Nooruddin2003SimplificationAR}, semantic labels are then assigned to each voxel based on part annotations. For shape part ground truth, a part dataset is generated by enlarging and centering all shape parts in the $32^3$ volumetric space. The inverse transformation matrices ground truth is also computed and saved in this step. The dataset is split into the training set and the test set with a ratio of $80\% / 20\%$. We use the chair category as the main demonstration category in this paper. See more results on other categories in the supplementary material.

In the autoencoder, we use convolution kernels of $4\times4\times4$ like \cite{Wu2016LearningAP,Li2020LearningPG} to avoid unnecessary output padding for deconvolution layers. Given input 3D shapes with a resolution of $32^3$, our encoder consists of four 3D convolution layers with kernel sizes of $4\times4\times4$ and strides of 2. Batch normalization and LeakyReLU layers are inserted between convolution layers. After latent representation decomposition, a shared decoder is applied to all part representations. The decoder simply mirrors the encoder except that a \textit{Sigmoid} nonlinearity is used in the last layer. The final outputs are the decoded parts that have been enlarged and centered in the volumetric space with a resolution of $32^3$.

%In the followed step of learning transformation matrices, the autoencoder weights trained in step 1 are frozen.
Based on the defined decoder, we have six feature layers that may be used. We number the latent representation as feature layer 0, the decoded parts as feature layer 5, and the feature maps in each decoder layer as feature layer 1-4. For example, feature layer 3 has a feature map size of $8^3$ with a channel number of 128 in our case. More details are given in the encoder-decoder architecture table in the supplementary materials. Note that layer 0 and layer 1 are different when channel-wise part attention is applied. Layer 0 has a feature size of 256 with 1 feature channel, while layer 1 has a feature size of $1^3$ with 256 feature channels. In the actual experiments, layer 1 and layer 2 are of too large feature channel size thus consuming extremely large GPU memory when the channel-wise strategy is applied. Hence they are less used in the following experiments.

Figure \ref{fig:block} gives a detailed network design of an attention block. It consists of an 8-heads attention network and a feed-forward network. Figure \ref{fig:net} gives the detailed network design of our proposed attention models. Both normal part attention and channel-wise part attention models are presented. Optional attention consistency loss $L_\text{AC}$ may be applied between the latent representations of different feature layers after the attention network (green frame box).

\begin{figure}[t]
\centering
\resizebox{1.0\linewidth}{!}{
\begin{tabular}{p{1.4cm}<{\centering} p{1.55cm}<{\centering} p{1.55cm}<{\centering} c p{1.55cm}<{\centering} p{1.55cm}<{\centering} c p{1.75cm}<{\centering} p{1.75cm}<{\centering}}
\multirow{2}{*}{\ Input \ } & \multicolumn{2}{c}{Simple MLP} &  & \multicolumn{2}{c}{Part Attention} &  & \multicolumn{2}{c}{Channel-wise Part Attention} \\ \cmidrule{2-3} \cmidrule{5-6} \cmidrule{8-9} 
 & step 2 & step 3 &  & step 2 & step 3 &  & step 2 & step 3 \\ \midrule
\end{tabular}}
\includegraphics[width=1.0\linewidth,trim=2 2 2 2,clip]{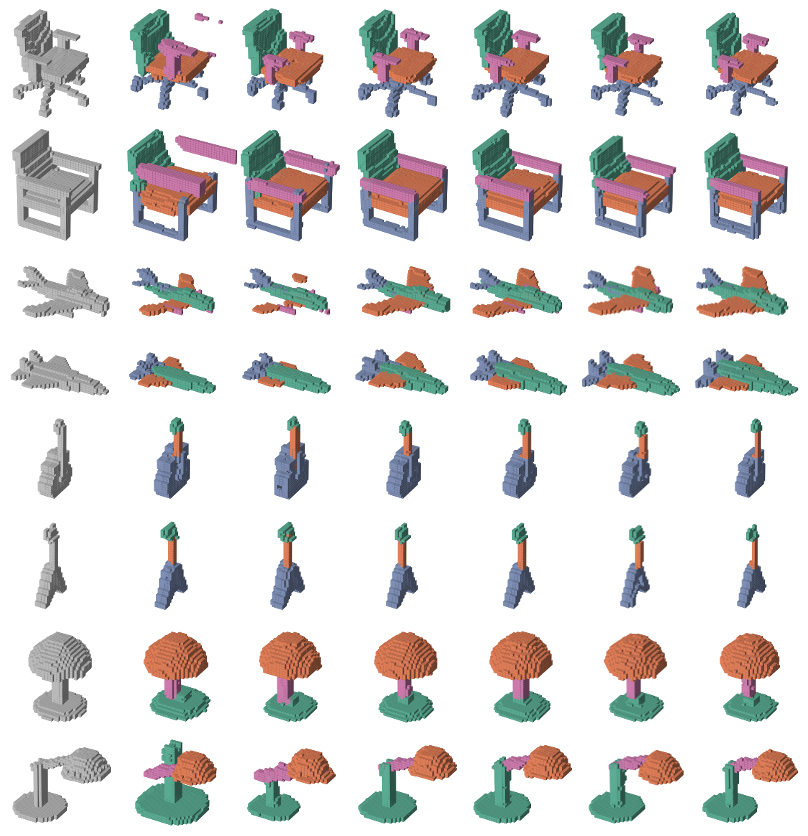}
\caption{Reconstruction results of our attention-based methods in comparison with applying simple MLP. Results before and after the fine-tuning are both presented. \vspace{-0.cm}}
\label{fig:vis_recon}
\end{figure}

\subsection{Training Parameters}
In step 1, we set $\gamma = 0.6$ for $L_\text{part}$. The weights for loss terms are set as $\omega_\text{PI}=\omega_\text{part}=1$. We use Adam optimizer with an initial learning rate of 0.001, the decay ratio is 0.8 with a decay step of every 50 epochs. The whole step 1 training takes 250 epochs.

In step 2, training settings are different for the baseline case and the attention-based case. Attention-based networks are usually more delicate and need more time to converge. For the baseline result of applying simple MLP, we use a learning rate of 0.001 with a decay ratio of 0.8 and a decay step of every 100 epochs. The training takes 500 epochs. 
For attention-based methods, the weights for loss terms are set as $\omega_\text{trans}=\omega_\text{AC}=1$ if $L_\text{AC}$ is applied. We use a learning rate of 0.0001 without decay and a training epochs number of 1500. The feature embedding dimension $d_A=256$. Like Transformers \cite{Vaswani2017AttentionIA}, a multi-head attention structure with 8 heads is used in each attention block.

For step 3, the initial learning rate has to be very small for fine-tuning since the final learning rate in step 1 is already of a very small number. Here, we use a learning rate of 0.00001 with a decay step of 250 epochs. The total training takes 500 epochs. $\gamma = 0.6$ is used for the full shape binary cross entropy loss $L_\text{shape}$. For the loss weights, we set $\omega_\text{PI}=1, \omega_\text{part}=1, \omega_\text{trans}= 10, \omega_\text{shape}= 10$, and $\omega_\text{AC}=1$ if it is applied. See the supplementary materials for an ablation study on hyper-parameter selection.

\subsection{Qualitative and Quantitative Results}

\begin{table}[t]
\centering
%\scalebox{0.85}{
\resizebox{1.0\linewidth}{!}{
\begin{tabular}{ccccccccc}
\toprule
\multirow{2}{*}{Model} & \multirow{2}{*}{\ Step \ } & \multicolumn{5}{c}{Part mIoU $\uparrow$} & 
\multirow{2}{*}{ Trans MSE $\downarrow$}
& \multirow{2}{*}{Shape mIoU $\uparrow$} \\ \cmidrule{3-7}
 &  & Back & Seat & Leg & Armrest & Mean &  &  \\ \midrule
% \multirow{2}{*}{CompSM \cite{Dubrovina2019CompositeSM}} & 1,2 & 65.4\% & 66.5\% & 64.6\% & 60.8\% & 65.9\% & 45.5 & 61.7\% $\pm$ 1.2\% \\
%  & 3 & 64.3\% & 65.9\% & 63.7\% & 60.2\% & 65.3\% & 47.2 & 64.4\% \\ \cmidrule{2-9} 
\multirow{2}{*}{Simple MLP} & 1,2 & 71.6\% & 73.2\% & 70.1\% & 61.2\% & 72.8\% & 38.5 & 63.8\% $\pm$ 0.1\% \\
 & 3 & 71.5\% & 72.8\% & 69.8\% & 60.4\% & 72.5\% & 42.6 & 71.8\% \\ \cmidrule{2-9} 
\multirow{2}{*}{Part Attention} & 1,2 & 71.6\% & 73.2\% & 70.1\% & 61.2\% & 72.8\% & 32.0 & 70.9\% $\pm$ 2.9\% \\
 & 3 & 71.6\% & 73.2\% & 70.1\% & 61.3\% & 72.8\% & \textbf{31.4} & 76.5\% \\ \cmidrule{2-9} 
\multirow{2}{*}{\begin{tabular}[c]{@{}c@{}}Channel-wise\\  Part Attention\end{tabular}} & 1,2 & 71.6\% & 73.2\% & 70.1\% & 61.2\% & 72.8\% & 36.9 & 75.8\% $\pm$ 2.1\% \\
 & 3 & 71.7\% & 73.2\% & 70.1\% & 61.3\% & 72.8\% & 36.2 & \textbf{78.4\%} \\ \bottomrule
\end{tabular}}
\caption{Quantitative results from different network models on the chair category. Note that for more precise comparison, we use an encoder of the same weights from step 1 for our models, thus their part mIoU before fine-tuning are identical.}
\label{table:miou}
\end{table}

\begin{table}[t]
\centering
%\scalebox{0.85}{
\resizebox{0.75\linewidth}{!}{
\begin{tabular}{cccc}
\toprule
\multirow{2}{*}{Model} & \multicolumn{3}{c}{Shape mIoU $\uparrow$} \\ \cmidrule{2-4} 
& Chair & Airplane & Lamp \\ \midrule
Pix2Mesh \cite{Wang2018Pixel2MeshG3} & 39.6\% & 42.0\% & 32.3\% \\
3DCNN \cite{Brock2016GenerativeAD} & 52.6\% & 50.6\% & 36.2\% \\
3D-R2N2 \cite{Choy20163DR2N2AU} & 55.0\% & 56.1\% & 42.1\% \\
OccNet \cite{Mescheder2019OccupancyNL} & 50.1\% & 57.1\% & 37.1\% \\
IM-Net \cite{Chen2019LearningIF} & 52.2\% & 55.4\% & 29.6\% \\
D2IM-Net \cite{Li2021D2IMNetLD} & 56.1\% & 55.8\% & 42.1\% \\ 
PQ-Net \cite{Wu2020PQNETAG} & 67.3\% & - & 41.3\% \\
CompSM \cite{Dubrovina2019CompositeSM} & 64.4\% & - & - \\ \midrule
Simple MLP & 71.8\% & 71.6\% & 65.4\% \\ 
Ours (Part Attention) & 76.5\% & \textbf{78.2\%} & 71.6\% \\
Ours (C-wise Part Attention) & \textbf{78.4\%} & 74.7\% & \textbf{72.6\%} \\ \bottomrule
\end{tabular}}
\caption{Quantitative shape reconstruction results compared to other methods. Evaluated with the shape mIoU metric on multiple categories.}
\label{table:miou-all}
\end{table}

Figure \ref{fig:vis_recon} gives some shape reconstruction results from unlabeled test 3D shapes. Parts are firstly generated and subsequently assembled with the transformation matrices learned by the attention network. Baseline results of using simple MLP are also given. Results with or without the fine-tuning in step 3 are both given.
From the visualization results, we can observe that applying simple MLP directly produces modest results yet the shape parts are wrongly scaled and connected in detail. It even fails to correctly reconstruct small-volume parts in some cases. 
On the other hand, our attention-based methods, both normal part attention and channel-wise part attention, produce more part-coherent reconstruction results on all categories, even for small-volume parts. More visualization results are given in the supplementary materials.

Quantitative results on metrics of part mIoU, shape mIoU, and transformation MSE are presented in Table \ref{table:miou}. All the evaluations are performed on the test dataset. Both results before and after the fine-tuning are presented. Larger mIoU means better reconstruction results for both parts and full shape, while smaller transformation MSE means better transformation matrices are learned. 
From Table \ref{table:miou}, we can clearly observe that compared to applying simple MLP, our attention-based method gets better numerical results on both transformation MSE and shape mIoU metrics in both steps. The simple MLP method even sacrifices part mIoU and transformation MSE performance for better shape mIoU performance in step 3. The normal part attention model gets the best transformation MSE performance, which means the visualized assembly results from it are more structurally correct. And the channel-wise part attention model gets the best shape mIoU performance, which means it focuses more on the overall reconstruction. This is actually consistent with our qualitative results.

The quantitative comparison results with other methods on the shape mIoU metric and the symmetry metric are presented in Table \ref{table:miou-all} and Table \ref{table:symmetry} respectively. 
The symmetry score is defined as the percentage of the shape voxels that got matched with their reflection given a vertical symmetry plane. Table \ref{table:symmetry} shows that our method is capable of generating more symmetric parts even for small volume parts like chair legs or armrests.

\begin{table}[t]
\centering
\resizebox{0.9\linewidth}{!}{
\begin{tabular}{c p{0.8cm}<{\centering} p{0.8cm}<{\centering} p{0.8cm}<{\centering} p{1.0cm}<{\centering} p{1.5cm}<{\centering}}
\toprule
\multirow{2}{*}{Model} & \multicolumn{5}{c}{Symmetry score $\uparrow$} \\ \cmidrule{2-6}
 & back & seat & leg & armrest & full shape  \\ \midrule
GT & 0.91 & 0.95 & 0.85 & 0.81 & 0.96  \\ \midrule
3D-GAN \cite{Wu2016LearningAP} & 0.71 & 0.76 & 0.40 & 0.16 & 0.70  \\
G2L-GAN \cite{Wang2018GlobaltolocalGM} & \textbf{0.93} & \textbf{0.94} & 0.74 & 0.64 & \textbf{0.91} \\
PAGENet \cite{Li2020LearningPG} & 0.88 & 0.90 & 0.68 & 0.64 & 0.85 \\ \midrule
Simple MLP  & 0.91 & \textbf{0.94} & 0.83 & 0.79 & 0.89 \\ 
Ours (Part Attention) & 0.92 & \textbf{0.94} & \textbf{0.84} & \textbf{0.82} & \textbf{0.91} \\
Ours (C-wise Part Attention) & 0.92 & \textbf{0.94} & \textbf{0.84} & \textbf{0.82} & 0.90  \\ \bottomrule
\end{tabular}}
\caption{Symmetry score of our proposed method compared to others on the chair category. Note that it is possible that the symmetry score is better than the ground truth.}
\label{table:symmetry}
\end{table}

Additionally, to quantitatively compare with methods that work on other 3D data representations, \eg, primitive shape-based and point clouds, we follow a similar scheme in SAGNet \cite{Wu2019StructureawareGN} to firstly convert 3D volumetric data to point clouds and then use the metrics proposed in \cite{Achlioptas2018LearningRA}. In Table \ref{table:mmd}, the results are presented using several metrics for 3D shape sets, including Jensen-Shannon Divergence (JSD), Coverage (COV), and Minimum Matching Distance (MMD). The latter two metrics are calculated using both the Chamfer Distance (CD) and Earth Mover’s Distance (EMD) for measuring the distance between shapes.

\begin{table}[t]
\centering
%\scalebox{0.85}{
\resizebox{1.0\linewidth}{!}{
\begin{tabular}{c p{1.2cm}<{\centering} cccc}
\toprule
Model & JSD $\downarrow$  & MMD-CD $\downarrow$ & MMD-EMD $\downarrow$ & COV-CD $\uparrow$ & COV-EMD $\uparrow$ \\ \midrule
G2L \cite{Wang2018GlobaltolocalGM} & 0.0357 & 0.0034 & 0.0682 & 0.837 & 0.834 \\
SAGNet \cite{Wu2019StructureawareGN} & 0.0342 & 0.0024 & 0.0608 & 0.751 & 0.743 \\
GRASS \cite{Li2017GRASSGR} & 0.0374 & 0.0030 & 0.0744 & 0.460 & 0.445 \\ 
StructureNet \cite{Mo2019StructureNetHG} & 0.0477 & 0.0097 & 0.1524 & 0.297 & 0.317 \\
BAE-Net \cite{Chen2019BAENETBA} & 0.7380 & 0.0057 & 0.3210 & 0.090 & 0.050 \\
MRGAN \cite{Gal2021MRGANM3} & 0.2460 & 0.0021 & 0.1660 & 0.670 & 0.230 \\
\textsc{Edit}VAE \cite{Li2021EditVAEUP} & 0.0310 & 0.0017 & 0.1010 & 0.450 & 0.390 \\ \midrule
Ours (Part Attention) & \textbf{0.0292} & 0.0018 & \textbf{0.0412} & \textbf{1.000} & \textbf{1.000} \\
Ours (C-wise Part Attention) & 0.0295 & \textbf{0.0016} & 0.0451 & \textbf{1.000} & \textbf{1.000} \\
\bottomrule
\end{tabular}}
\caption{Quantitative evaluation for shape modeling on the chair category. For JSD and MMD, the smaller the better. For COV, the larger the better. Since our method is reconstruction-based, we achieve 1.0 on COV metric.}
\label{table:mmd}
\end{table}

All the above results from the attention-based models are using feature layers 0/3/5 as the input. For the normal part attention mode, $L_\text{AC}$ is applied; for the channel-wise part attention mode, $L_\text{AC}$ is not applied. See an ablation study in subsection \ref{subsec:attentionAblation} for reasons of using this choice.

\subsection{Learned Part Relations}

%-------------- Figure: attention map -----------------
\begin{figure}[t]
    \centering
    \begin{subfigure}[b]{0.8\linewidth}
        \centering
        \includegraphics[width=\linewidth,trim=2 2 2 2,clip]{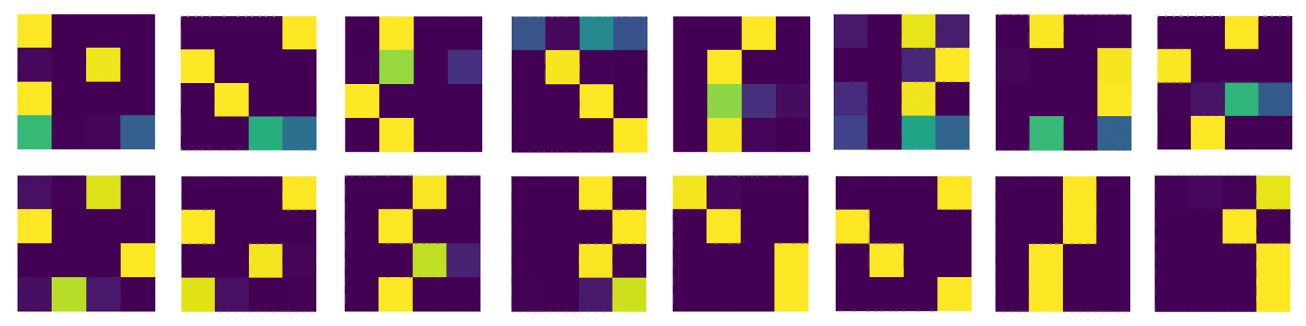}
        \caption{Attention maps in the first attention block. \vspace{0.1cm}}
        \label{fig:attenMap_block1}
    \end{subfigure}
    \quad\quad
    \begin{subfigure}[b]{0.8\linewidth}
        \centering
        \includegraphics[width=\linewidth,trim=2 2 2 2,clip]{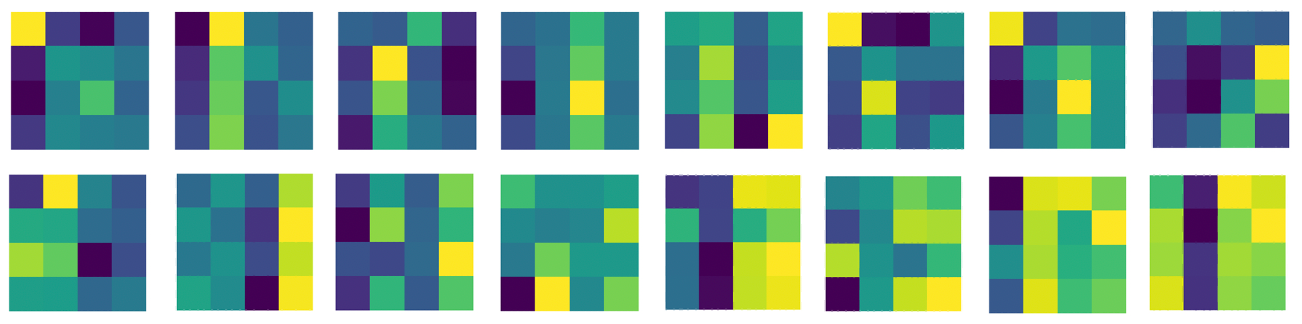}
        \caption{Attention maps in the last attention block.\vspace{-0.cm}}
        \label{fig:attenMap_block3}
    \end{subfigure}
    \caption{Learned attention maps in different attention blocks. Lighter yellow means higher correlation, darker blue means lower correlation.\vspace{-0.cm}}
    \label{fig:attenMap}
\end{figure}

%-------------- Figure: swap -----------------
\begin{figure}[t]
    \centering
    \begin{subfigure}[b]{0.8\linewidth}
        \centering
        \resizebox{1.0\linewidth}{!}{
        \begin{tabular}{p{1.4cm}<{\centering} p{1.55cm}<{\centering} p{1.55cm}<{\centering} c p{1.55cm}<{\centering} p{1.55cm}<{\centering}}
        \multirow{2}{*}{\ GT \ } & \multicolumn{2}{c}{Simple MLP} &  & \multicolumn{2}{c}{Attention-based} \\ \cmidrule{2-3} \cmidrule{5-6}
        & Recon & Swap &  & Recon & Swap \\ \midrule
        \end{tabular}}
        %\vspace{0.1cm}
        \includegraphics[width=\linewidth,trim=2 2 2 2,clip]{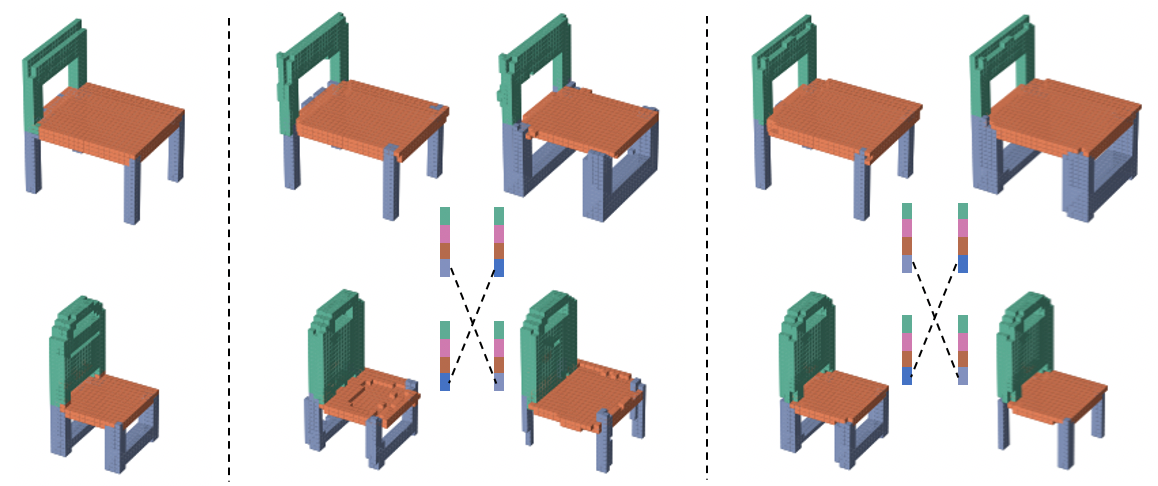}
        \caption{Chair leg swapping}
        \label{fig:vis_swap_a}
    \end{subfigure}
    \quad\quad
    \begin{subfigure}[b]{0.8\linewidth}
        \centering
        \includegraphics[width=\linewidth,trim=2 2 2 2,clip]{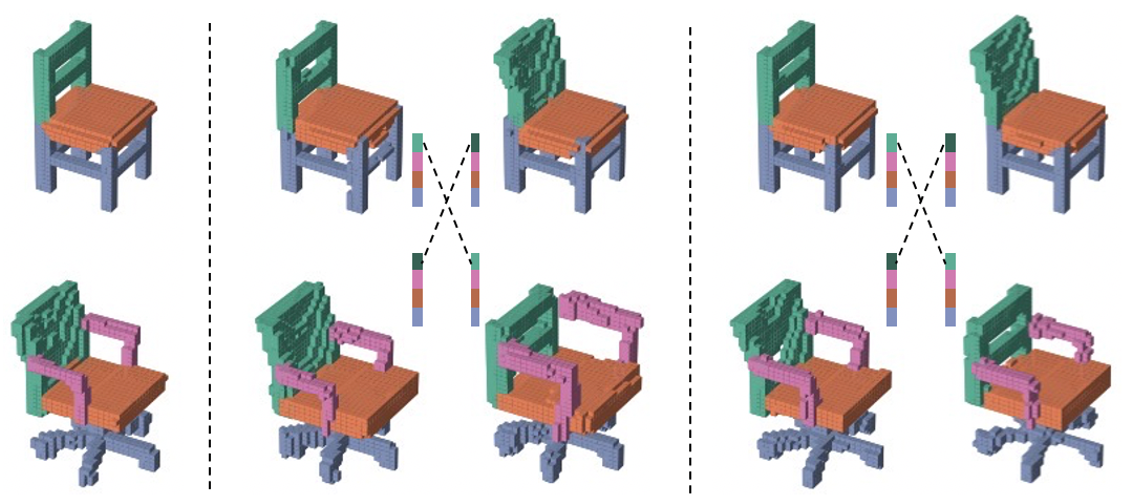}
        \caption{Chair back swapping \vspace{-0.cm}}
        \label{fig:vis_swap_b}
    \end{subfigure}
    \caption{Part assembly results after swapping certain part representations in the latent space. \vspace{-0.cm}}
    \label{fig:vis_swap}
\end{figure}

\textbf{Learned Attention Maps.}
The attention maps, i.e., the part correlation matrices, is the key to the success of our proposed method. Figure \ref{fig:attenMap_block1} and Figure \ref{fig:attenMap_block3} give some typical attention maps in different attention heads from the first attention block and the last attention block, respectively. Lighter yellow means higher part correlation, while darker blue means lower part correlation. From the figures, we can observe that in the first attention block the module mostly focuses on learning the correlation of one part to one part, or sometimes to two parts, in the attention heads (feature maps are mostly composed of light yellow or dark blue, only few greenish squares). Meanwhile in the attention heads from the last attention block, the module mostly focuses on learning the correlation of one part to three or four parts (feature maps are mostly composed of various greenish squares, only few dark blue squares). This means our network architecture learns more part correlation information going deeper through the attention blocks.

\textbf{Swap.} To demonstrate that our attention-based network is able to learn good part relations, we conducted experiments of swapping shape parts in the latent space. 
Figure \ref{fig:vis_swap_a} and Figure \ref{fig:vis_swap_b} give a demo of swapping the chair legs or the chair back respectively. The simple MLP method fails to transform the swapped parts correctly according to the size of other parts. Swapping one part may even cause incorrect changes in other parts, \eg, seat in Figure \ref{fig:vis_swap_a} and armrest in Figure \ref{fig:vis_swap_b}. Our attention-based method successfully overcomes all those problems. Swapped parts are correctly aligned with other parts without harming them.

\begin{figure}[t]
\centering
\includegraphics[width=0.8\linewidth,trim=2 2 2 2,clip]{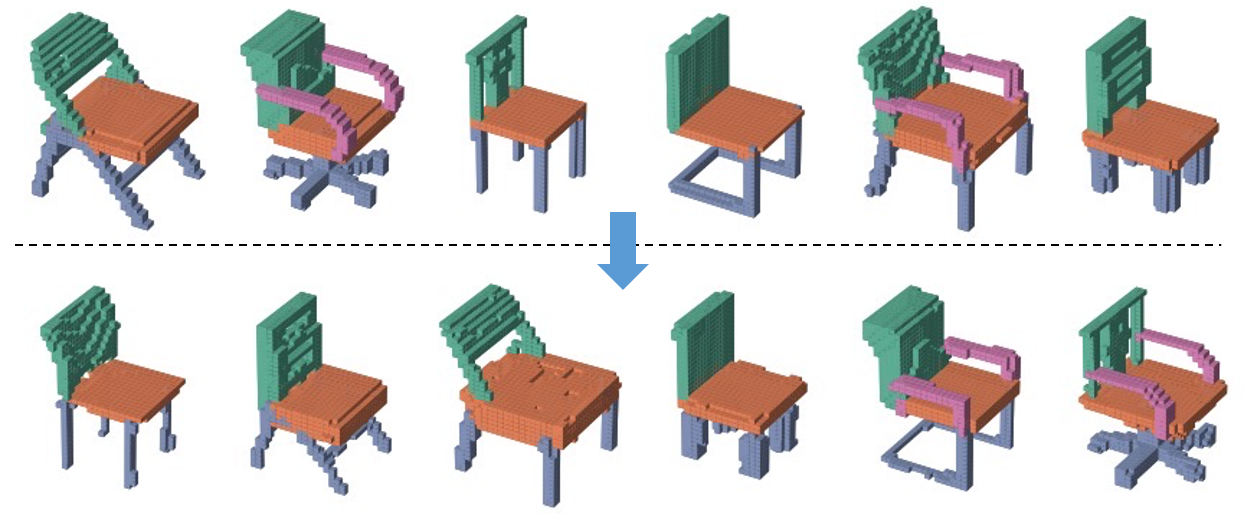}
\vspace{-0.cm}
\caption{Part assembly results of mixing random parts from different input shapes to generate new shapes.}
\label{fig:vis_mix}
\end{figure}

\textbf{Mix.} Figure \ref{fig:vis_mix} gives some part assembly results by using random parts from the shape category to compose new shapes. From the figure, we can observe that for the newly composed shapes, the transformation matrices of different parts are learned coherently for the assembly.

\begin{figure}[t]
\centering
\includegraphics[width=\linewidth,trim=2 2 2 2,clip]{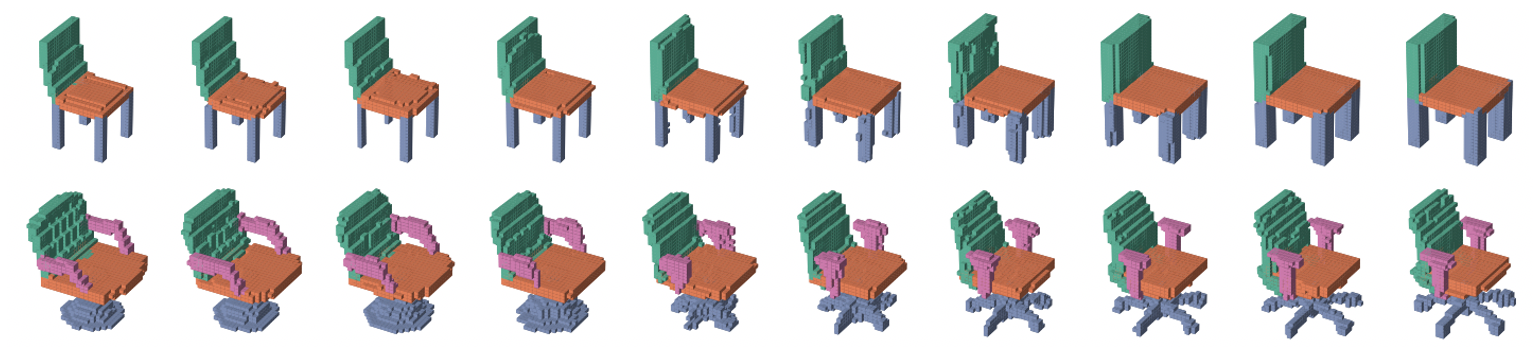}
\vspace{-0.5cm}
\caption{Interpolation results between chair pairs.\vspace{-0.cm}}
\label{fig:vis_interpolation}
\end{figure}

\textbf{Interpolation.} Additionally, we also give a demo of interpolating between chair pairs in the latent space in Figure \ref{fig:vis_interpolation}. The transition between the same parts in the pairs is smooth, \eg, part thickness, or the morphing chair legs.

%-------------- Figure: mix -----------------

%-------------- Figure: interpolation -----------------

\subsection{Ablation Study on Attention Models}
\label{subsec:attentionAblation}
After conducting a more detailed ablation study on the VoxAttention models with different architecture choices, in Table \ref{table:attention}, we report quantitative results of using different feature layers or combinations of them as input for both normal part attention case and channel-wise part attention case. For settings with multiple feature layers used as attention input, results with or without the additional $L_\text{AC}$ are both given.

In step 2, the full shape reconstruction loss is not applied. Hence for the shape mIoU metric, a certain fluctuation (max. $\pm$ 5\%) in the performance is observed in most cases. This is actually quite reasonable: a small difference in transformation matrices may cause a big difference in shape mIoU. For example, if the transformed chair back is moved one voxel length ahead, its transformation matrix will not make big difference, but the shape mIoU will change a lot since the shape parts are usually only several voxels thick. In this case, in Table \ref{table:attention}, instead of smoothing the curve with a large smoothing parameter, we give a mean value of shape mIoU with a standard deviation using the numerical results in the last 100 epochs.

\begin{table}[t]
\centering
%\scalebox{0.85}{
\resizebox{1.0\linewidth}{!}{
\begin{tabular}{ccccccc}
\toprule
\multirow{2}{*}{Feature layer(s)} & \multirow{2}{*}{\begin{tabular}[c]{@{}c@{}}\ Apply \ \\ $L_\text{AC}$ \end{tabular}} & \multicolumn{2}{c}{Part Attention} &  & \multicolumn{2}{c}{Channel-wise Part Attention} \\ \cmidrule{3-4} \cmidrule{6-7} 
 &  & trans MSE $\downarrow$ & shape mIoU $\uparrow$ &  & trans MSE $\downarrow$ & shape mIoU $\uparrow$ \\ \midrule
0 & - & \textbf{31.5} & 70.5\% $\pm$ 4.4\% &  & \textbf{31.8} & 69.5\% $\pm$ 3.6\% \\
1 & - & \textbf{31.5} & 70.4\% $\pm$ 3.1\% &  & 88.4 & 53.2\% $\pm$ 3.1\% \\
2 & - & 32.3 & \textbf{72.9\% $\pm$ 1.7\%} &  & 37.9 & 70.3\% $\pm$ 0.8\% \\
3 & - & 33.0 & 71.1\% $\pm$ 2.2\% &  & 35.2 & \textbf{73.3\% $\pm$ 1.5\%} \\
4 & - & 37.3 & 65.9\% $\pm$ 2.3\% &  & 33.9 & 67.2\% $\pm$ 3.2\% \\
5 & - & 42.1 & 66.4\% $\pm$ 2.6\% &  & 41.9 & 62.5\% $\pm$ 5.1\% \\ \midrule
\multirow{2}{*}{02} & no & 39.3 & 64.5\% $\pm$ 5.1\% &  & 55.4 & 55.7\% $\pm$ 2.2\% \\
 & yes & 32.2 & 70.0\% $\pm$ 2.2\% &  & 33.5 & 71.7\% $\pm$ 1.2\% \\ \cmidrule{2-7} 
\multirow{2}{*}{03} & no & 60.8 & 39.4\% $\pm$ 3.6\% &  & 35.0 & 69.9\% $\pm$ 3.3\% \\
 & yes & 32.4 & 69.0\% $\pm$ 4.7\% &  & 33.3 & 68.1\% $\pm$ 2.3\% \\ \cmidrule{2-7} 
\multirow{2}{*}{05} & no & 354.1 & 16.9\% $\pm$ 1.6\% &  & 34.9 & 70.2\% $\pm$ 2.9\% \\
 & yes & 33.3 & 68.0\% $\pm$ 3.8\% &  & 35.2 & 69.7\% $\pm$ 2.8\% \\ \cmidrule{2-7} 
\multirow{2}{*}{023} & no & 99.4 & 35.6\% $\pm$ 1.7\% &  & 76.3 & 45.7\% $\pm$ 4.2\% \\
 & yes & 32.0 & 68.8\% $\pm$ 2.9\% &  & 34.5 & 69.2\% $\pm$ 1.9\% \\ \cmidrule{2-7} 
\multirow{2}{*}{035} & no & 232.0 & 24.5\% $\pm$ 1.2\% &  & 36.9 & \textbf{75.8\% $\pm$ 2.1\%} \\ 
 & yes & 32.0 & \textbf{70.9\% $\pm$ 2.9\%} &  & 33.2 & 70.1\% $\pm$ 2.1\% \\ \cmidrule{2-7} 
\multirow{2}{*}{045} & no & 211.9 & 15.0\% $\pm$ 1.7\% &  & 34.8 & 68.8\% $\pm$ 3.2\% \\
 & yes & 32.6 & 69.4\% $\pm$ 1.8\% &  & 32.1 & 70.6\% $\pm$ 2.3\% \\ \cmidrule{2-7} 
\multirow{2}{*}{0345} & no & 286.3 & 24.9\% $\pm$ 1.0\% &  & 91.9 & 31.8\% $\pm$ 3.2\% \\
 & yes & 31.3 & 70.4\% $\pm$ 2.8\% &  & \textbf{31.3} & 71.7\% $\pm$ 1.7\% \\ \cmidrule{2-7} 
\multirow{2}{*}{012345} & no & 390.7 & 22.7\% $\pm$ 1.2\% &  & - & - \\
 & yes & \textbf{31.1} & 70.4\% $\pm$ 3.9\% &  & - & - \\ \toprule
\end{tabular}}
\caption{Quantitative evaluation results in step 2 on the chair category with different experimental settings. Numbers in the first column are the layer index of feature layers used as input.}
\label{table:attention}
\end{table}

\noindent The following conclusions can be drawn from Table \ref{table:attention}:
\begin{itemize}[itemsep=-2pt,topsep=0pt,left=3pt]
\item For the settings that are identical to the network, their performances are similar to each other. For example, normal part attention with input layer 0, normal part attention with input layer 1, and channel-wise part attention with input layer 0.
\item When a single feature layer is used as input, for normal part attention models, using front feature layers gets a smaller transformation MSE. However, for channel-wise part attention models, using rear-end feature layers may get a smaller transformation MSE since the network needs to balance the trade-off between feature size and feature channel number.
\item When multiple feature layers are used as input in normal part attention models, $L_\text{AC}$ must be applied. Otherwise, the model collapses with a large transformation MSE and a bad shape mIoU.
\item When multiple feature layers are used as input in the channel-wise part attention models, $L_\text{AC}$ is not necessary anymore when only one middle feature layer (layer index 1-4) is used. 
\item Transformation MSE indicates the shape mIoU to some degree. But they do not have an absolute strong correlation. Also, better shape mIoU does not ensure better visualization results since the full shape reconstruction loss is not introduced in this step. Multiple layer input usually yields better visualization results in shape details compared to single layer input.
\end{itemize}

Balancing the trade-off between model size, metrics results, and visualization qualities, we use feature layers 0/3/5 as the input for most results in the previous subsections. With this choice, for the normal part attention mode, $L_\text{AC}$ is applied; for the channel-wise part attention mode, $L_\text{AC}$ is not applied.
More intuitive evaluation curves along the training are given in the supplementary material.

\section{Conclusions}
\label{sec:conclusion}
In this paper, a novel attention-based part assembly method has been proposed for 3D shape modeling. Both qualitative and quantitative results demonstrate that the proposed method achieves better performance on this task compared to other state-of-the-art methods. The channel-wise strategy and the additional attention consistency loss also contribute to the good results. 
For future possible topics, we would like to try patch-based self-attention to see if the semantic information and the part relations can be learned in a self-supervised manner. Applying attention-based methods on other 3D data representations, \eg, point clouds or meshes, to learn semantic information would also be an interesting research direction.

%%%%%%%%% REFERENCES
{\small
\bibliographystyle{ieee_fullname}
\bibliography{main}

\begin{thebibliography}{10}\itemsep=-1pt

\bibitem{Achlioptas2018LearningRA}
Panos Achlioptas, Olga Diamanti, Ioannis Mitliagkas, and Leonidas~J. Guibas.
\newblock Learning representations and generative models for 3d point clouds.
\newblock In {\em ICML}, 2018.

\bibitem{Balashova2018StructureAwareSS}
Elena Balashova, Vivek Singh, Jiangping Wang, Brian Teixeira, Terrence Chen,
  and Thomas~A. Funkhouser.
\newblock Structure-aware shape synthesis.
\newblock {\em 2018 International Conference on 3D Vision (3DV)}, pages
  140--149, 2018.

\bibitem{Brock2016GenerativeAD}
Andrew Brock, Theodore Lim, James~M. Ritchie, and Nick Weston.
\newblock Generative and discriminative voxel modeling with convolutional
  neural networks.
\newblock {\em ArXiv}, abs/1608.04236, 2016.

\bibitem{Carion2020EndtoEndOD}
Nicolas Carion, Francisco Massa, Gabriel Synnaeve, Nicolas Usunier, Alexander
  Kirillov, and Sergey Zagoruyko.
\newblock End-to-end object detection with transformers.
\newblock {\em ArXiv}, abs/2005.12872, 2020.

\bibitem{Chang2015ShapeNetAI}
Angel~X. Chang, Thomas~A. Funkhouser, Leonidas~J. Guibas, Pat Hanrahan, Qixing
  Huang, Zimo Li, Silvio Savarese, Manolis Savva, Shuran Song, Hao Su,
  Jianxiong Xiao, L. Yi, and Fisher Yu.
\newblock Shapenet: An information-rich 3d model repository.
\newblock {\em ArXiv}, abs/1512.03012, 2015.

\bibitem{Chen2021PreTrainedIP}
Hanting Chen, Yunhe Wang, Tianyu Guo, Chang Xu, Yiping Deng, Zhenhua Liu, Siwei
  Ma, Chunjing Xu, Chao Xu, and Wen Gao.
\newblock Pre-trained image processing transformer.
\newblock {\em 2021 IEEE/CVF Conference on Computer Vision and Pattern
  Recognition (CVPR)}, pages 12294--12305, 2021.

\bibitem{Chen2020GenerativePF}
Mark Chen, Alec Radford, Jeff Wu, Heewoo Jun, Prafulla Dhariwal, David Luan,
  and Ilya Sutskever.
\newblock Generative pretraining from pixels.
\newblock In {\em ICML}, 2020.

\bibitem{Chen2019BAENETBA}
Zhiqin Chen, K. Yin, Matthew Fisher, Siddhartha Chaudhuri, and Hao Zhang.
\newblock Bae-net: Branched autoencoder for shape co-segmentation.
\newblock {\em 2019 IEEE/CVF International Conference on Computer Vision
  (ICCV)}, pages 8489--8498, 2019.

\bibitem{Chen2019LearningIF}
Zhiqin Chen and Hao Zhang.
\newblock Learning implicit fields for generative shape modeling.
\newblock {\em 2019 IEEE/CVF Conference on Computer Vision and Pattern
  Recognition (CVPR)}, pages 5932--5941, 2019.

\bibitem{Choy20163DR2N2AU}
Christopher~Bongsoo Choy, Danfei Xu, JunYoung Gwak, Kevin Chen, and Silvio
  Savarese.
\newblock 3d-r2n2: A unified approach for single and multi-view 3d object
  reconstruction.
\newblock In {\em ECCV}, 2016.

\bibitem{Dosovitskiy2021AnII}
Alexey Dosovitskiy, Lucas Beyer, Alexander Kolesnikov, Dirk Weissenborn,
  Xiaohua Zhai, Thomas Unterthiner, Mostafa Dehghani, Matthias Minderer, Georg
  Heigold, Sylvain Gelly, Jakob Uszkoreit, and Neil Houlsby.
\newblock An image is worth 16x16 words: Transformers for image recognition at
  scale.
\newblock {\em ArXiv}, abs/2010.11929, 2021.

\bibitem{Dubrovina2019CompositeSM}
Anastasia Dubrovina, F. Xia, Panos Achlioptas, Mira Shalah, and Leonidas~J.
  Guibas.
\newblock Composite shape modeling via latent space factorization.
\newblock {\em 2019 IEEE/CVF International Conference on Computer Vision
  (ICCV)}, pages 8139--8148, 2019.

\bibitem{Engel2021PointT}
Nico Engel, Vasileios Belagiannis, and Klaus C.~J. Dietmayer.
\newblock Point transformer.
\newblock {\em IEEE Access}, 9:134826--134840, 2021.

\bibitem{Gal2021MRGANM3}
Rinon Gal, Amit~H. Bermano, Hao Zhang, and Daniel Cohen-Or.
\newblock Mrgan: Multi-rooted 3d shape representation learning with
  unsupervised part disentanglement.
\newblock {\em 2021 IEEE/CVF International Conference on Computer Vision
  Workshops (ICCVW)}, pages 2039--2048, 2021.

\bibitem{Girdhar2016LearningAP}
Rohit Girdhar, David~F. Fouhey, Mikel~D. Rodriguez, and Abhinav~Kumar Gupta.
\newblock Learning a predictable and generative vector representation for
  objects.
\newblock {\em ArXiv}, abs/1603.08637, 2016.

\bibitem{Guo2021PCTPC}
Meng-Hao Guo, Junxiong Cai, Zheng-Ning Liu, Tai-Jiang Mu, Ralph~Robert Martin,
  and Shimin Hu.
\newblock Pct: Point cloud transformer.
\newblock {\em Comput. Vis. Media}, 7:187--199, 2021.

\bibitem{Han2020ASO}
Kai Han, Yunhe Wang, Hanting Chen, Xinghao Chen, Jianyuan Guo, Zhenhua Liu,
  Yehui Tang, An Xiao, Chunjing Xu, Yixing Xu, Zhaohui Yang, Yiman Zhang, and
  Dacheng Tao.
\newblock A survey on visual transformer.
\newblock {\em ArXiv}, abs/2012.12556, 2020.

\bibitem{He2020SVGANetSV}
Qingdong He, Zhengning Wang, Hao Zeng, Yi Zeng, Shuaicheng Liu, and Bing Zeng.
\newblock Svga-net: Sparse voxel-graph attention network for 3d object
  detection from point clouds.
\newblock {\em ArXiv}, abs/2006.04043, 2020.

\bibitem{Henderson2020Leveraging2D}
Paul Henderson, Vagia Tsiminaki, and Christoph~H. Lampert.
\newblock Leveraging 2d data to learn textured 3d mesh generation.
\newblock {\em 2020 IEEE/CVF Conference on Computer Vision and Pattern
  Recognition (CVPR)}, pages 7495--7504, 2020.

\bibitem{Li2020LearningPG}
Jun Li, Chengjie Niu, and Kai Xu.
\newblock Learning part generation and assembly for structure-aware shape
  synthesis.
\newblock In {\em AAAI}, 2020.

\bibitem{Li2017GRASSGR}
Jun Li, Kai Xu, Siddhartha Chaudhuri, Ersin Yumer, Hao Zhang, and Leonidas~J.
  Guibas.
\newblock Grass: Generative recursive autoencoders for shape structures.
\newblock {\em ArXiv}, abs/1705.02090, 2017.

\bibitem{Li2021D2IMNetLD}
Manyi Li and Hao Zhang.
\newblock D2im-net: Learning detail disentangled implicit fields from single
  images.
\newblock {\em 2021 IEEE/CVF Conference on Computer Vision and Pattern
  Recognition (CVPR)}, pages 10241--10250, 2021.

\bibitem{Li2021EditVAEUP}
Shidi Li, Miaomiao Liu, and Christian~J. Walder.
\newblock Editvae: Unsupervised part-aware controllable 3d point cloud shape
  generation.
\newblock {\em ArXiv}, abs/2110.06679, 2021.

\bibitem{Mao2021VoxelTF}
Jiageng Mao, Yujing Xue, Minzhe Niu, Haoyue Bai, Jiashi Feng, Xiaodan Liang,
  Hang Xu, and Chunjing Xu.
\newblock Voxel transformer for 3d object detection.
\newblock {\em 2021 IEEE/CVF International Conference on Computer Vision
  (ICCV)}, pages 3144--3153, 2021.

\bibitem{Mescheder2019OccupancyNL}
Lars~M. Mescheder, Michael Oechsle, Michael Niemeyer, Sebastian Nowozin, and
  Andreas Geiger.
\newblock Occupancy networks: Learning 3d reconstruction in function space.
\newblock {\em 2019 IEEE/CVF Conference on Computer Vision and Pattern
  Recognition (CVPR)}, pages 4455--4465, 2019.

\bibitem{Mo2019StructureNetHG}
Kaichun Mo, Paul Guerrero, L. Yi, Hao Su, Peter Wonka, Niloy~Jyoti Mitra, and
  Leonidas~J. Guibas.
\newblock Structurenet: Hierarchical graph networks for 3d shape generation.
\newblock {\em ACM Trans. Graph.}, 38:242:1--242:19, 2019.

\bibitem{Mo2020StructEditLS}
Kaichun Mo, Paul Guerrero, L. Yi, Hao Su, Peter Wonka, Niloy~Jyoti Mitra, and
  Leonidas~J. Guibas.
\newblock Structedit: Learning structural shape variations.
\newblock {\em 2020 IEEE/CVF Conference on Computer Vision and Pattern
  Recognition (CVPR)}, pages 8856--8865, 2020.

\bibitem{Nooruddin2003SimplificationAR}
Fakir~S. Nooruddin and Greg Turk.
\newblock Simplification and repair of polygonal models using volumetric
  techniques.
\newblock {\em IEEE Trans. Vis. Comput. Graph.}, 9:191--205, 2003.

\bibitem{Park2019DeepSDFLC}
Jeong~Joon Park, Peter~R. Florence, Julian Straub, Richard~A. Newcombe, and S.
  Lovegrove.
\newblock Deepsdf: Learning continuous signed distance functions for shape
  representation.
\newblock {\em 2019 IEEE/CVF Conference on Computer Vision and Pattern
  Recognition (CVPR)}, pages 165--174, 2019.

\bibitem{Pumarola2020CFlowCG}
Albert Pumarola, Stefan Popov, Francesc Moreno-Noguer, and Vittorio Ferrari.
\newblock C-flow: Conditional generative flow models for images and 3d point
  clouds.
\newblock {\em 2020 IEEE/CVF Conference on Computer Vision and Pattern
  Recognition (CVPR)}, pages 7946--7955, 2020.

\bibitem{Riegler2017OctNetLD}
Gernot Riegler, Ali~O. Ulusoy, and Andreas Geiger.
\newblock Octnet: Learning deep 3d representations at high resolutions.
\newblock {\em 2017 IEEE Conference on Computer Vision and Pattern Recognition
  (CVPR)}, pages 6620--6629, 2017.

\bibitem{Schor2019CompoNetLT}
Nadav Schor, Oren Katzir, Hao Zhang, and Daniel Cohen-Or.
\newblock Componet: Learning to generate the unseen by part synthesis and
  composition.
\newblock {\em 2019 IEEE/CVF International Conference on Computer Vision
  (ICCV)}, pages 8758--8767, 2019.

\bibitem{Tulsiani2018MultiviewCA}
Shubham Tulsiani, Alexei~A. Efros, and Jitendra Malik.
\newblock Multi-view consistency as supervisory signal for learning shape and
  pose prediction.
\newblock {\em 2018 IEEE/CVF Conference on Computer Vision and Pattern
  Recognition}, pages 2897--2905, 2018.

\bibitem{Valsesia2019LearningLG}
Diego Valsesia, Giulia Fracastoro, and Enrico Magli.
\newblock Learning localized generative models for 3d point clouds via graph
  convolution.
\newblock In {\em ICLR}, 2019.

\bibitem{Vaswani2017AttentionIA}
Ashish Vaswani, Noam~M. Shazeer, Niki Parmar, Jakob Uszkoreit, Llion Jones,
  Aidan~N. Gomez, Lukasz Kaiser, and Illia Polosukhin.
\newblock Attention is all you need.
\newblock {\em ArXiv}, abs/1706.03762, 2017.

\bibitem{Wang2018GlobaltolocalGM}
H. Wang, Nadav Schor, Ruizhen Hu, Haibin Huang, Daniel Cohen-Or, and Hui Huang.
\newblock Global-to-local generative model for 3d shapes.
\newblock {\em ACM Transactions on Graphics (TOG)}, 37:1 -- 10, 2018.

\bibitem{Wang2018Pixel2MeshG3}
Nanyang Wang, Yinda Zhang, Zhuwen Li, Yanwei Fu, W. Liu, and Yu-Gang Jiang.
\newblock Pixel2mesh: Generating 3d mesh models from single rgb images.
\newblock {\em ArXiv}, abs/1804.01654, 2018.

\bibitem{wang2017cnn}
Peng-Shuai Wang, Yang Liu, Yu-Xiao Guo, Chun-Yu Sun, and Xin Tong.
\newblock O-cnn: Octree-based convolutional neural networks for 3d shape
  analysis.
\newblock {\em ACM Transactions On Graphics (TOG)}, 36(4):1--11, 2017.

\bibitem{Wen2019Pixel2MeshM3}
Chao Wen, Yinda Zhang, Zhuwen Li, and Yanwei Fu.
\newblock Pixel2mesh++: Multi-view 3d mesh generation via deformation.
\newblock {\em 2019 IEEE/CVF International Conference on Computer Vision
  (ICCV)}, pages 1042--1051, 2019.

\bibitem{Wu2022SwitchVAE}
Chengzhi Wu, Julius Pfrommer, Mingyuan Zhou, and Jürgen Beyerer.
\newblock Generative-contrastive learning for self-supervised latent
  representations of 3d shapes from multi-modal euclidean input.
\newblock {\em 36th Conference on Artificial Intelligence AAAI Workshop}, 2022.

\bibitem{Wu2016LearningAP}
Jiajun Wu, Chengkai Zhang, Tianfan Xue, Bill Freeman, and Joshua~B. Tenenbaum.
\newblock Learning a probabilistic latent space of object shapes via 3d
  generative-adversarial modeling.
\newblock In {\em NIPS}, 2016.

\bibitem{Wu2020PQNETAG}
Rundi Wu, Yixin Zhuang, Kai Xu, Hao Zhang, and Baoquan Chen.
\newblock Pq-net: A generative part seq2seq network for 3d shapes.
\newblock {\em 2020 IEEE/CVF Conference on Computer Vision and Pattern
  Recognition (CVPR)}, pages 826--835, 2020.

\bibitem{Wu20153DSA}
Zhirong Wu, Shuran Song, Aditya Khosla, Fisher Yu, Linguang Zhang, Xiaoou Tang,
  and Jianxiong Xiao.
\newblock 3d shapenets: A deep representation for volumetric shapes.
\newblock {\em 2015 IEEE Conference on Computer Vision and Pattern Recognition
  (CVPR)}, pages 1912--1920, 2015.

\bibitem{Wu2019StructureawareGN}
Zhijie Wu, Xiang Wang, Di Lin, Dani Lischinski, Daniel Cohen-Or, and Hui Huang.
\newblock Structure-aware generative network for 3d-shape modeling.
\newblock {\em ACM Trans. Graph.}, 38:91:1--91:14, 2019.

\bibitem{Yang2019PointFlow3P}
Guandao Yang, Xun Huang, Zekun Hao, Ming-Yu Liu, Serge~J. Belongie, and Bharath
  Hariharan.
\newblock Pointflow: 3d point cloud generation with continuous normalizing
  flows.
\newblock {\em 2019 IEEE/CVF International Conference on Computer Vision
  (ICCV)}, pages 4540--4549, 2019.

\bibitem{Yi2016ASA}
L. Yi, Vladimir~G. Kim, Duygu Ceylan, I-Chao Shen, Mengyan Yan, Hao Su, Cewu
  Lu, Qixing Huang, Alla Sheffer, and Leonidas~J. Guibas.
\newblock A scalable active framework for region annotation in 3d shape
  collections.
\newblock {\em ACM Transactions on Graphics (TOG)}, 35:1 -- 12, 2016.

\bibitem{Yin2020COALESCECA}
K. Yin, Zhiqin Chen, Siddhartha Chaudhuri, Matthew Fisher, Vladimir~G. Kim, and
  Hao Zhang.
\newblock Coalesce: Component assembly by learning to synthesize connections.
\newblock {\em 2020 International Conference on 3D Vision (3DV)}, pages 61--70,
  2020.

\bibitem{Zhao2021PointT}
Hengshuang Zhao, Li Jiang, Jiaya Jia, Philip H.~S. Torr, and Vladlen Koltun.
\newblock Point transformer.
\newblock {\em 2021 IEEE/CVF International Conference on Computer Vision
  (ICCV)}, pages 16239--16248, 2021.

\bibitem{Zhu2018SCORESSC}
Chenyang Zhu, Kai Xu, Siddhartha Chaudhuri, Renjiao Yi, and Hao Zhang.
\newblock Scores: Shape composition with recursive substructure priors.
\newblock {\em ArXiv}, abs/1809.05398, 2018.

\bibitem{Zou20173DPRNNGS}
Chuhang Zou, Ersin Yumer, Jimei Yang, Duygu Ceylan, and Derek Hoiem.
\newblock 3d-prnn: Generating shape primitives with recurrent neural networks.
\newblock {\em 2017 IEEE International Conference on Computer Vision (ICCV)},
  pages 900--909, 2017.

\end{thebibliography}
}

\end{document}

% If your paper is accepted and the title of your paper is very long,
% the style will print as headings an error message. Use the following
% command to supply a shorter title of your paper so that it can be
% used as headings.
%
%\runningtitle{I use this title instead because the last one was very long}

% If your paper is accepted and the number of authors is large, the
% style will print as headings an error message. Use the following
% command to supply a shorter version of the authors names so that
% they can be used as headings (for example, use only the surnames)
%
%\runningauthor{Surname 1, Surname 2, Surname 3, ...., Surname n}

% Supplementary material: To improve readability, you must use a single-column format for the supplementary material.
\onecolumn
\aistatstitle{Attention-based Part Assembly for 3D Volumetric Shape Modeling \\ - Supplementary Materials}

\aistatsauthor{\large Chengzhi Wu$^{1}$ \quad Junwei Zheng$^{1}$ \quad Julius Pfrommer$^{2,3}$ \quad Jürgen Beyerer$^{1,2,3}$}
\vspace{2pt}
\aistatsaddress{\large $^{1}$Karlsruhe Institute of Technology, Germany \quad
$^{2}$Fraunhofer IOSB, Germany\\
\large $^{3}$Fraunhofer Center for Machine Learning, Germany \vspace{-0.3cm}}

\section{Network Architecture}
\subsection{Encoder and Decoder}
The detailed network structures of the encoder and decoder are defined as follows. The accordingly feature layer indices are also given. 

\begin{table}[h]
\centering
\begin{tabular}{p{1.6cm}<{\centering} p{1.6cm}<{\centering} p{1.6cm}<{\centering} ccc}
\toprule
Layer & Kernel & Stride & Activation & \ Channel \ & Output shape \\ \midrule
conv. & $4 \times 4 \times 4$ & $2 \times 2 \times 2$ & LeakyReLU & 64 & $16^3$ \\
conv. & $4 \times 4 \times 4$ & $2 \times 2 \times 2$ & LeakyReLU & 128 & $8^3$ \\
conv. & $4 \times 4 \times 4$ & $2 \times 2 \times 2$ & LeakyReLU & 256 & $4^3$ \\
conv. & $4 \times 4 \times 4$ & $1 \times 1 \times 1$ & LeakyReLU & 256 & $1^3$ \\
reshape & - & - & - & 1 & 256 \\ \bottomrule 
\end{tabular}
\caption{Encoder Structure.}
\label{table:encoder}
\end{table}

%\vspace{-1cm}

\begin{table}[h]
\centering
%\scalebox{0.9}{
\begin{tabular}{p{1.4cm}<{\centering} p{1.5cm}<{\centering} p{1.55cm}<{\centering} ccc|c}
\toprule
Layer & Kernel & Stride & Activation & \ Channel \ & Output shape \ & \begin{tabular}[c]{@{}c@{}} \ feature layer \\ index \end{tabular} \\ \midrule
input & - & - & - & 1 & 256 & 0 \\
reshape & - & - & - & 256 & $1^3$ & 1 \\
deconv. & $4 \times 4 \times 4$ & $1 \times 1 \times 1$ & LeakyReLU & 256 & $4^3$ & 2 \\
deconv. & $4 \times 4 \times 4$ & $2 \times 2 \times 2$ & LeakyReLU & 128 & $8^3$ & 3 \\
deconv. & $4 \times 4 \times 4$ & $2 \times 2 \times 2$ & LeakyReLU & 64 & $16^3$ & 4 \\
deconv. & $4 \times 4 \times 4$ & $2 \times 2 \times 2$ & Sigmoid & 1 & $32^3$ & 5 \\ \bottomrule
\end{tabular}
\caption{Decoder Structure.}
\label{table:decoder}
\end{table}

\iffalse
\subsection{Attention Block and Attention Network}
Figure \ref{fig:block} gives a detailed network design of an attention block. It consists of an 8-heads attention network and a feed-forward network. Figure \ref{fig:net} gives the detailed network design of our proposed attention models. Both normal part attention and channel-wise part attention models are presented. Optional attention consistency loss $L_\text{AC}$ may be applied between the latent representations of different feature layers after the attention network (green frame box).

\begin{figure}[h]
\centering
\includegraphics[width=\textwidth,trim=2 2 2 2,clip]{images_supp/block.png}
\caption{Detailed network design of an attention block.}
\label{fig:block}
\end{figure}

\begin{figure}[h]
\centering
\includegraphics[width=0.95\textwidth,trim=2 2 2 2,clip]{images_supp/net.png}
\caption{Detailed network design of proposed attention network architecture. Both normal part attention and channel-wise part attention models are presented. $N_p$ stands for the number of parts in the input category. $N_x$ stands for the total number of input feature layers.}
\label{fig:net}
\end{figure}
\fi

%\clearpage

\subsection{Number of Model Parameters}
The information is given in Table \ref{netParaNumber}. Compared to the simple MLP method, our proposed attention-based methods have less parameters. This is a common outcome when high-dimensional fully connected layers are replaced with attention blocks. Additionally, when the feature channel dimension is preserved, the model has even less parameters since the weights in the attention blocks are shared over all feature channels.

\begin{table}[h]
\centering
%\resizebox{0.8\linewidth}{!}{
\begin{tabular}{@{}cccc@{}}
\toprule
                     & Simple MLP \ & \ Part Attention \ & \ Channel-wise Part Attention \\ \midrule
Trainable parameters & 47.61M       & 46.25M           & 29.61M                    \\ \bottomrule
\end{tabular}%}
\caption{Number of parameters in different models.}
\label{netParaNumber}
\end{table}

\section{Loss Weights for Finetuning}
Table \ref{table:lossRatio} gives an ablation study of how we choose our loss weights for different loss terms in step 3. Here, we use a normal part attention model with input feature layers 0/3/5 and apply $L_\text{AC}$ as an example. Since the autoencoder is already well trained for part generation in step 1, we find that the change in part mIoU is negligible in the finetuning step. Meanwhile, using a larger $\omega_\text{trans}$ results in a smaller transformation MSE, and using a larger $\omega_\text{shape}$ results in a better shape mIoU. Balancing the trade-off between all those those numerical results and the visualization results, we use loss weights $\omega_\text{PI}=1, \omega_\text{part}=1, \omega_\text{trans}= 10, \omega_\text{shape}= 10$, and $\omega_\text{AC}=1$ if it is applied. \vspace{-0.0cm}

\begin{table}[h]
\centering
\begin{tabular}{ccccccccccccc}
\toprule
\multicolumn{5}{c}{Loss weights} & \multirow{2}{*}{} & \multicolumn{5}{c}{Part mIoU} & \multirow{2}{*}{\begin{tabular}[c]{@{}c@{}}\ Trans\ \\ \ MSE\ \end{tabular}} & \multirow{2}{*}{\ Shape mIoU} \\ \cmidrule{1-5} \cmidrule{7-11}
$\omega_\text{part}$ & $\omega_\text{PI}$ & $\omega_\text{trans}$ & $\omega_\text{AC}$ & $\omega_\text{shape}$ &  & back & seat & leg & armrest & mean &  &  \\ \midrule
1 & 1 & 1 & 1 & 1 &  & 71.6\% & 73.1\% & 70.1\% & 61.2\% & 72.8\% & 31.9 & 76.2\% \\
1 & 1 & 1 & 1 & 10 &  & 71.6\% & 73.2\% & 70.1\% & 61.2\% & 72.8\% & 31.8 & \textbf{76.5\%} \\
1 & 1 & 10 & 1 & 1 &  & 71.6\% & 73.1\% & 70.0\% & 61.2\% & 72.7\% & 31.5 & 76.3\% \\
1 & 1 & 10 & 1 & 10 &  & 71.6\% & 73.2\% & 70.1\% & 61.3\% & 72.8\% & \textbf{31.4} & \textbf{76.5\%} \\
0.1 & 0.1 & 1 & 1 & 1 &  & 71.6\% & 73.2\% & 70.1\% & 61.2\% & 72.8\% & 31.6 & 76.3\% \\ \bottomrule
\end{tabular}
\caption{The influence of using different loss weights choices. All models are based on an identical step 2 model.}
\label{table:lossRatio}
%\vspace{-0.5cm}
\end{table}

\section{Training Curves}
During the training, models are also evaluated with the test dataset every 10 epochs along the training. In our paper, an ablation study of the input feature layers choice for attention models is presented. In this supplementary material, their full shape mIoU and the transformation matrices MSE curves are given as follows for more insights. %\vspace{-0.5cm}

\begin{figure}[H]
\begin{minipage}{0.5\textwidth}
\centering
\includegraphics[width=\linewidth,trim=2 2 2 2,clip]{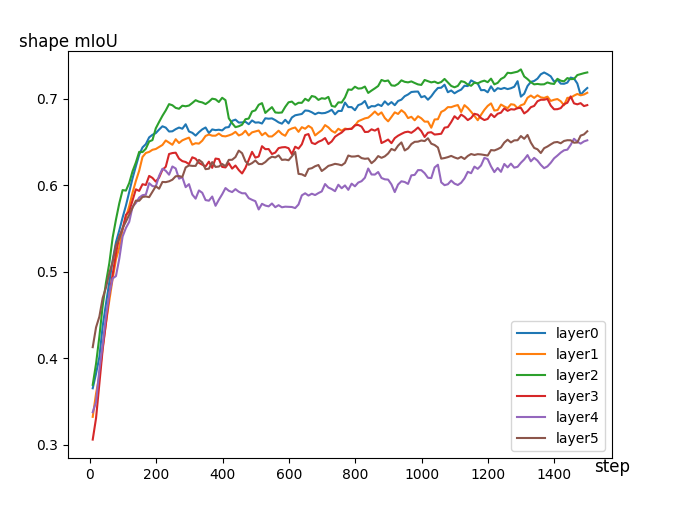}
\end{minipage}
\begin{minipage}{0.5\textwidth}
\centering
\includegraphics[width=\linewidth,trim=2 2 2 2,clip]{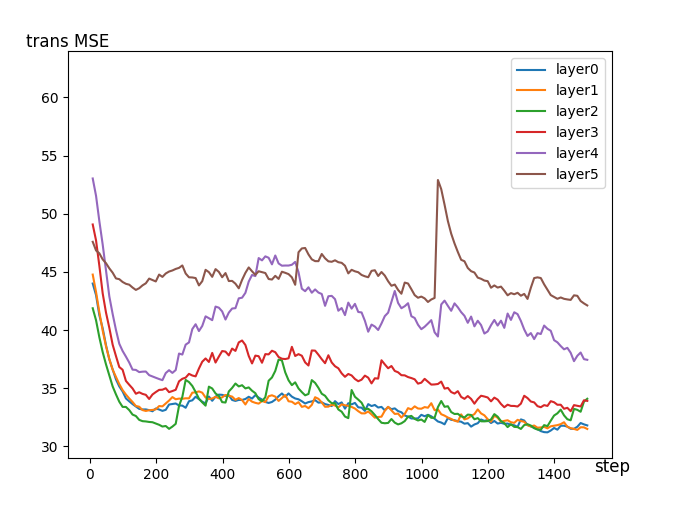}
\end{minipage}
\caption{Metric curves of using different single feature layer as input in a normal part attention model. \vspace{-0.cm}}
\label{fig:group1}
\end{figure}

\begin{figure}[H]
\begin{minipage}{0.5\textwidth}
\centering
\includegraphics[width=\linewidth,trim=2 2 2 2,clip]{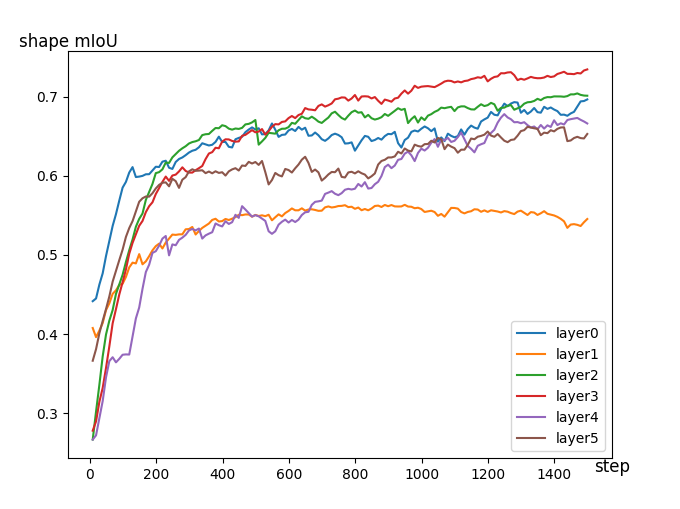}
\end{minipage}
\begin{minipage}{0.5\textwidth}
\centering
\includegraphics[width=\linewidth,trim=2 20 2 20,clip]{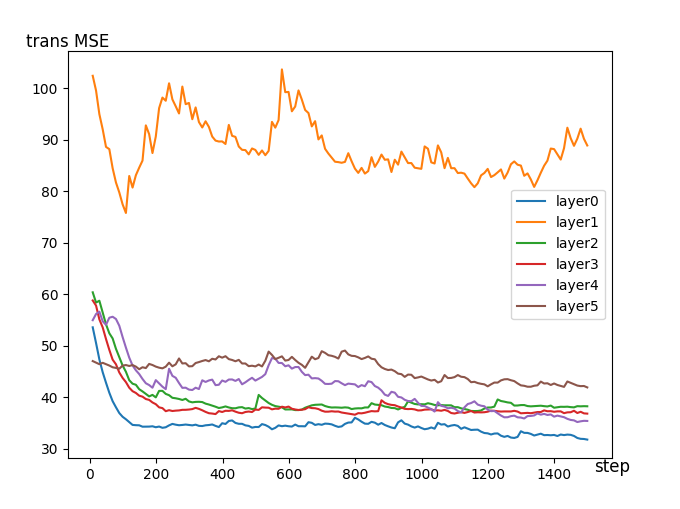}
\end{minipage}
\caption{Metric curves of using different single feature layer as input in a channel-wise part attention model. \vspace{-0.cm}}
\label{fig:group2}
\end{figure}

\begin{figure}[H]
\begin{minipage}{0.5\textwidth}
\centering
\includegraphics[width=\linewidth,trim=2 20 2 20,clip]{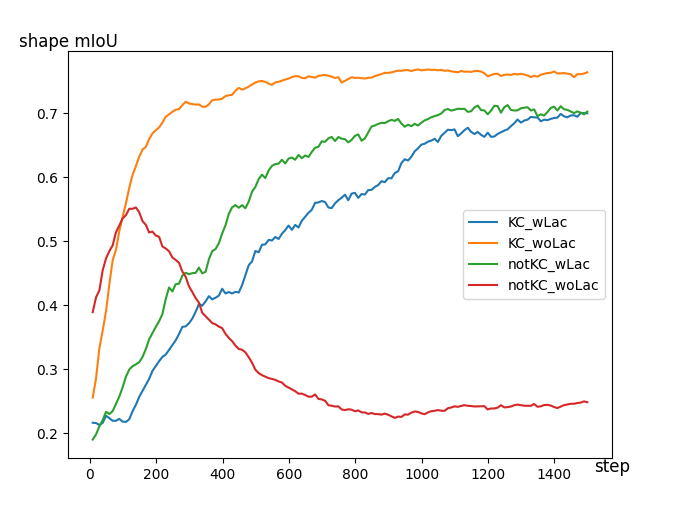}
\end{minipage}
\begin{minipage}{0.5\textwidth}
\centering
\includegraphics[width=\linewidth,trim=2 20 2 20,clip]{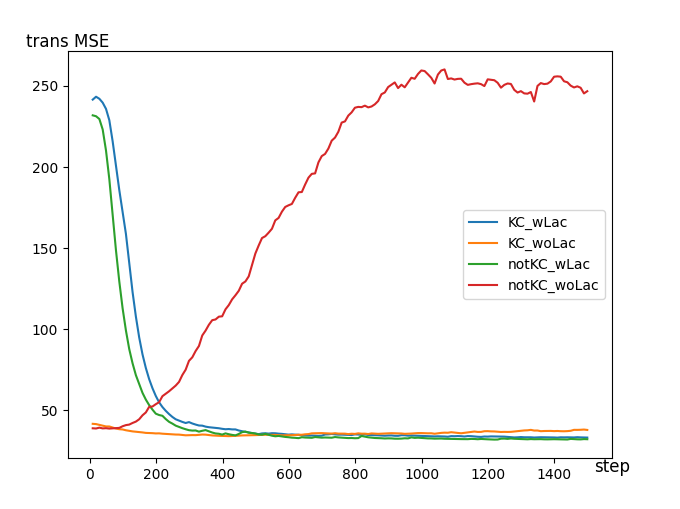}
\end{minipage}
\caption{Metric curves when using feature layers 0/3/5 as input. KC stands for using the keeping channel dimension strategy (channel-wise part attention), while notKC stands for the normal part attention mode. wLac stands for $L_\text{AC}$ is applied, while woLac stands for $L_\text{AC}$ is not applied.}
\label{fig:group3}
\end{figure}

Figure \ref{fig:group1} and figure \ref{fig:group2} give the metric curves of using different single feature layer as input in a normal part attention model or a channel-wise part attention model, respectively. The curves are in accordance with the conclusions we give in our main paper.
Additionally, under the condition of using feature layers 0/3/5 as input, the metric curves of using different network setting choices (attention mode, and the optional $L_\text{AC}$) are also given in figure \ref{fig:group3}. Based on those results, we finally choose the normal part attention model with $L_\text{AC}$, and the channel-wise part attention model without $L_\text{AC}$ as the main experimental settings.

\clearpage

\section{More Qualitative Results}
Results on the chair category are mainly presented in the paper. We hereby present more qualitative results on other categories including airplane, guitar, and lamp, on all kinds of tasks.

\subsection{Shape Reconstruction}
\begin{figure}[h]
\centering
\begin{tabular}{p{1.4cm}<{\centering} p{1.55cm}<{\centering} p{1.55cm}<{\centering} c p{1.55cm}<{\centering} p{1.55cm}<{\centering} c p{1.75cm}<{\centering} p{1.75cm}<{\centering}}
\multirow{2}{*}{\ Input \ } & \multicolumn{2}{c}{Simple MLP} &  & \multicolumn{2}{c}{Part Attention} &  & \multicolumn{2}{c}{Channel-wise Part Attention} \\ \cmidrule{2-3} \cmidrule{5-6} \cmidrule{8-9} 
 & step 2 & step 3 &  & step 2 & step 3 &  & step 2 & step 3 \\ \midrule
\end{tabular}
\includegraphics[width=0.85\textwidth,trim=2 2 2 2,clip]{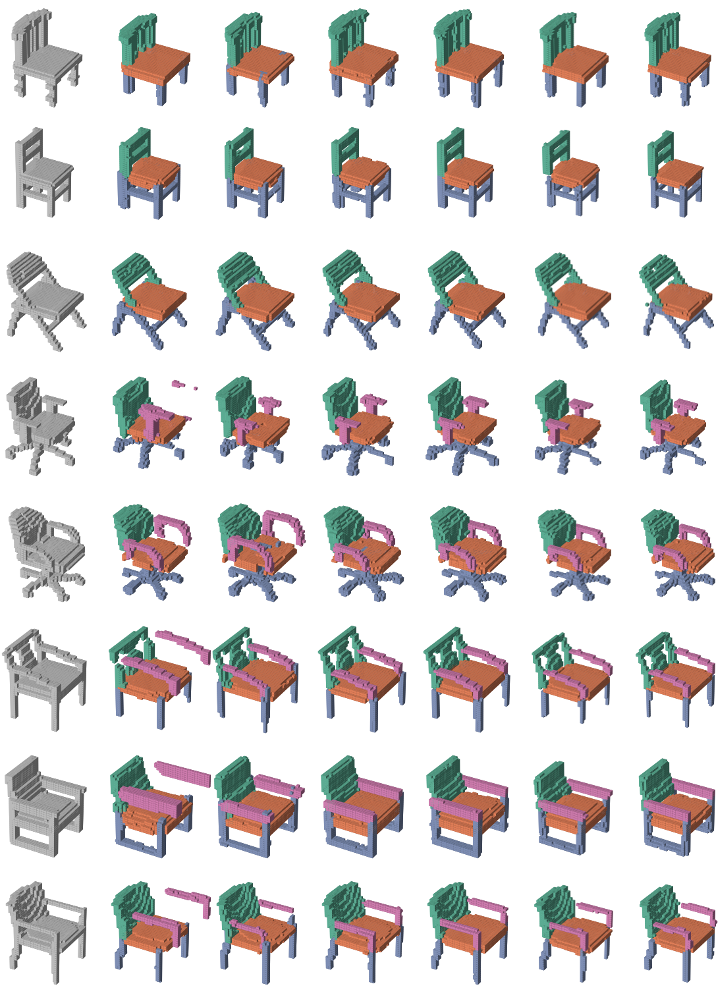}
\caption{Reconstruction results of our attention-based methods on the chair category in comparison of applying simple dense layers directly. Results before and after the finetuning are both presented.}
\label{fig:vis_recon1}
\end{figure}

\begin{figure}[h]
\centering
\begin{tabular}{p{1.4cm}<{\centering} p{1.55cm}<{\centering} p{1.55cm}<{\centering} c p{1.55cm}<{\centering} p{1.55cm}<{\centering} c p{1.75cm}<{\centering} p{1.75cm}<{\centering}}
\multirow{2}{*}{\ Input \ } & \multicolumn{2}{c}{Simple MLP} &  & \multicolumn{2}{c}{Part Attention} &  & \multicolumn{2}{c}{Channel-wise Part Attention} \\ \cmidrule{2-3} \cmidrule{5-6} \cmidrule{8-9} 
 & step 2 & step 3 &  & step 2 & step 3 &  & step 2 & step 3 \\ \midrule
\end{tabular}
\includegraphics[width=0.88\textwidth,trim=2 2 2 2,clip]{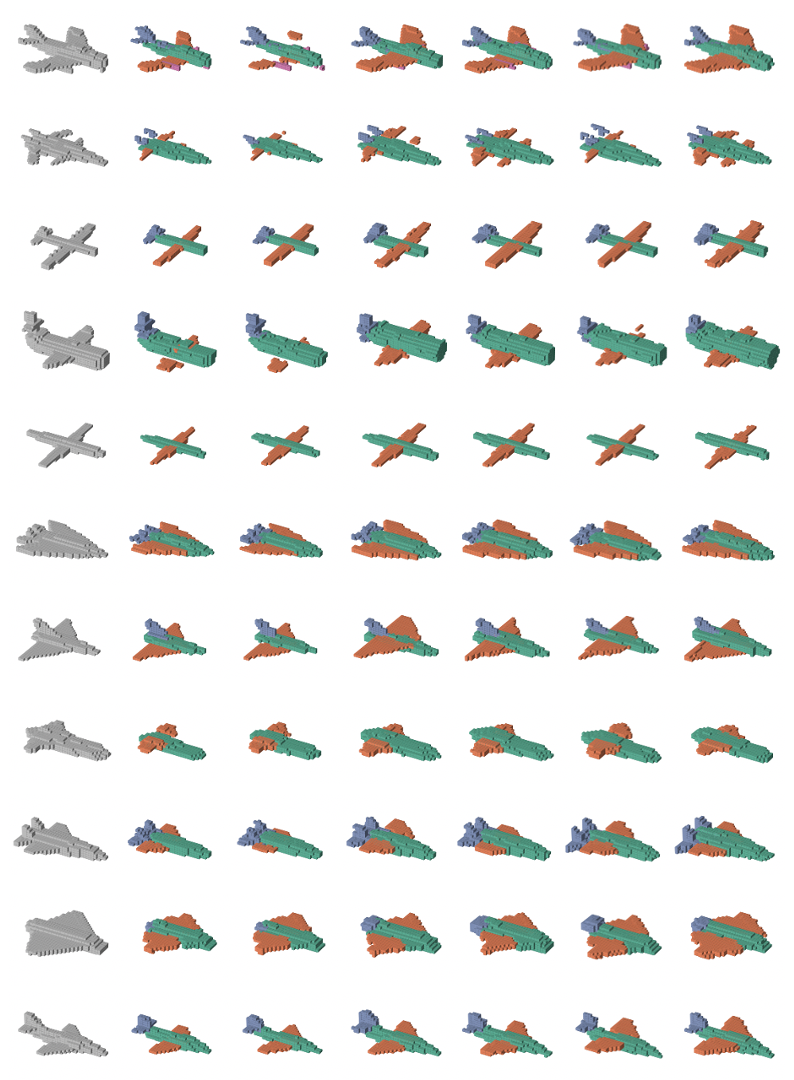}
\caption{Reconstruction results of our attention-based methods on the airplane category in comparison of applying simple dense layers directly. Results before and after the finetuning are both presented.}
\label{fig:vis_recon2}
\end{figure}

\begin{figure}[h]
\centering
\begin{tabular}{p{1.4cm}<{\centering} p{1.55cm}<{\centering} p{1.55cm}<{\centering} c p{1.55cm}<{\centering} p{1.55cm}<{\centering} c p{1.75cm}<{\centering} p{1.75cm}<{\centering}}
\multirow{2}{*}{\ Input \ } & \multicolumn{2}{c}{Simple MLP} &  & \multicolumn{2}{c}{Part Attention} &  & \multicolumn{2}{c}{Channel-wise Part Attention} \\ \cmidrule{2-3} \cmidrule{5-6} \cmidrule{8-9} 
 & step 2 & step 3 &  & step 2 & step 3 &  & step 2 & step 3 \\ \midrule
\end{tabular}
\includegraphics[width=0.85\textwidth,trim=2 2 2 2,clip]{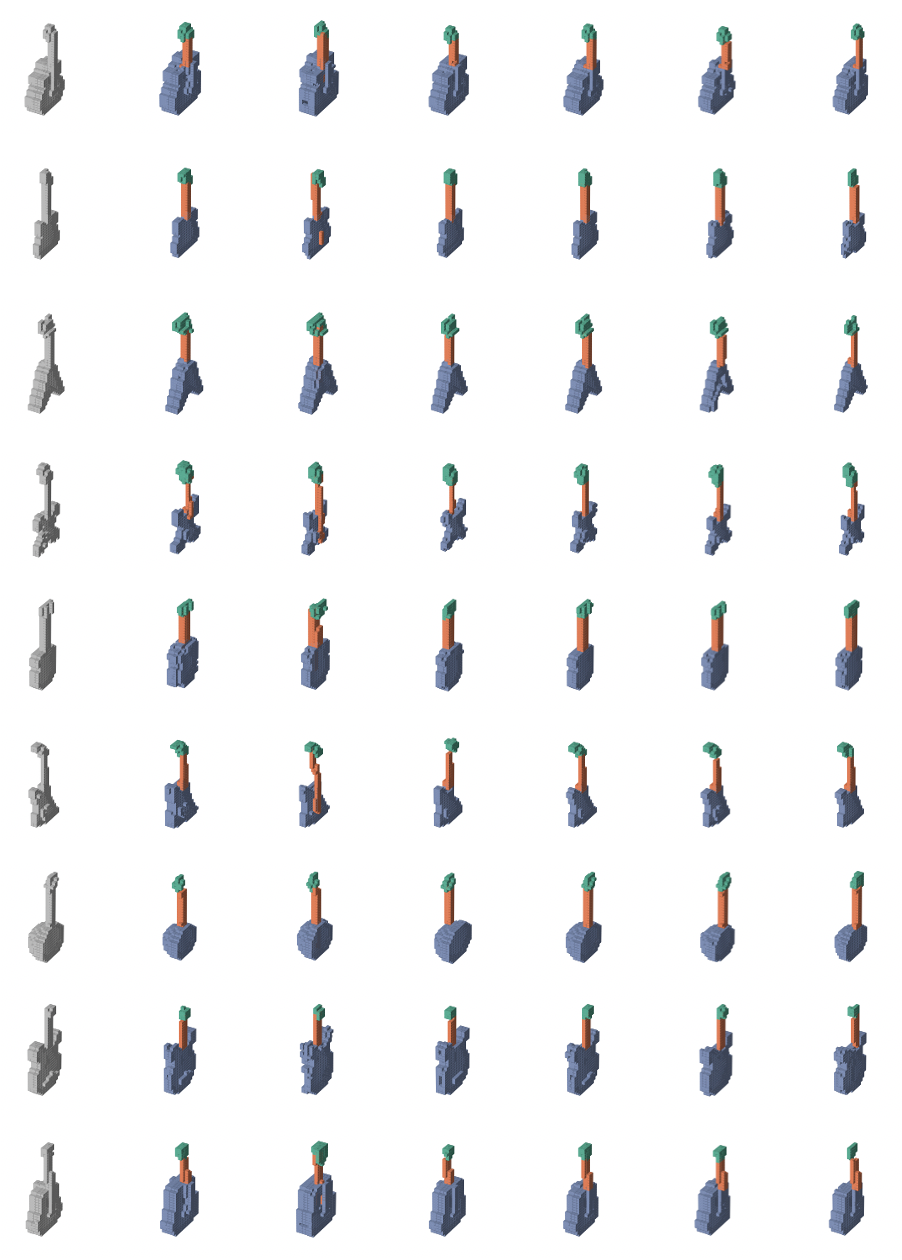}
\caption{Reconstruction results of our attention-based methods on the guitar category in comparison of applying simple dense layers directly. Results before and after the finetuning are both presented.}
\label{fig:vis_recon3}
\end{figure}

\begin{figure}[h]
\centering
\begin{tabular}{p{1.4cm}<{\centering} p{1.55cm}<{\centering} p{1.55cm}<{\centering} c p{1.55cm}<{\centering} p{1.55cm}<{\centering} c p{1.75cm}<{\centering} p{1.75cm}<{\centering}}
\multirow{2}{*}{\ Input \ } & \multicolumn{2}{c}{Simple MLP} &  & \multicolumn{2}{c}{Part Attention} &  & \multicolumn{2}{c}{Channel-wise Part Attention} \\ \cmidrule{2-3} \cmidrule{5-6} \cmidrule{8-9} 
 & step 2 & step 3 &  & step 2 & step 3 &  & step 2 & step 3 \\ \midrule
\end{tabular}
\includegraphics[width=0.88\textwidth,trim=2 2 2 2,clip]{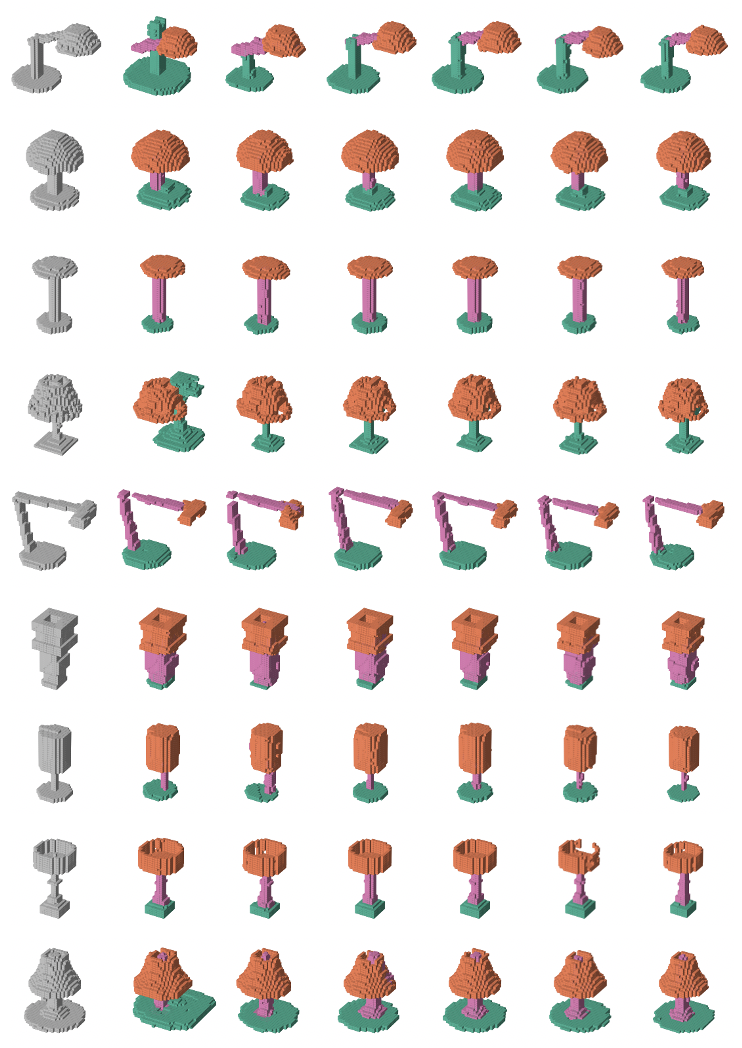}
\caption{Reconstruction results of our attention-based methods on the lamp category in comparison of applying simple dense layers directly. Results before and after the finetuning are both presented.}
\label{fig:vis_recon4}
\end{figure}

Following the same presentation way as in our paper,  Figures \ref{fig:vis_recon1}-\ref{fig:vis_recon4} give some additional reconstruction results on those categories respectively.

\clearpage

\clearpage

\clearpage

\clearpage

\clearpage
\subsection{Part Swapping}
\begin{figure}[H]
\centering
\begin{tabular}{ccccc}
Simple MLP \qquad \quad \qquad & \qquad Attention-based & \qquad \quad & Simple MLP \qquad & \qquad \qquad \quad Attention-based \\ \cmidrule{1-2} \cmidrule{4-5} 
\end{tabular}
\includegraphics[width=0.9\textwidth,trim=2 2 2 2,clip]{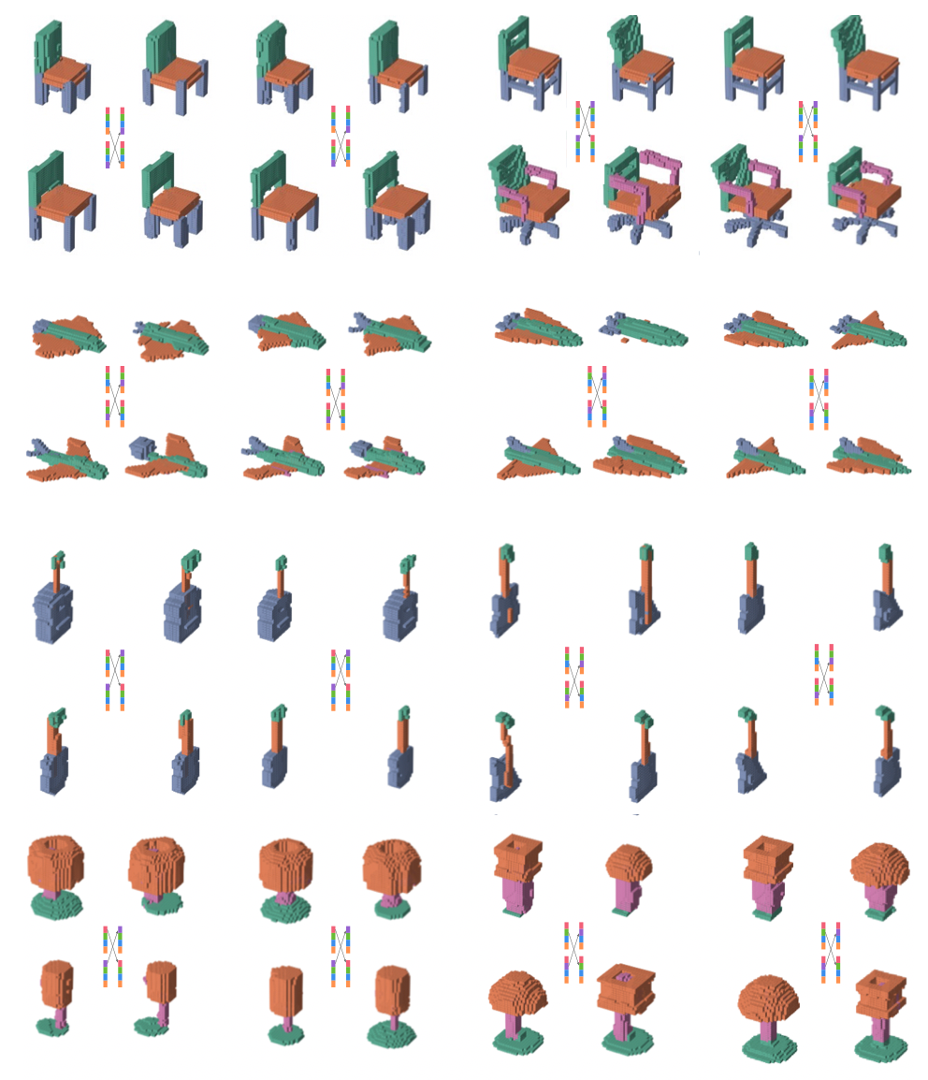}
\caption{Part swapping results of our attention-based methods in comparison to applying simple dense layers directly on all four categories.}
\label{fig:vis_swap}
\end{figure}

Figure \ref{fig:vis_swap} gives some experimental results of swapping shape parts in the latent space. From it, we can observe that compared to the simple MLP method, attention-based method performs much better in scaling and translating all shape parts to a relatively correct place when a part is swapped.

\clearpage
\subsection{Part Mixing for Random Assembly}
\begin{figure}[!h]
\begin{minipage}{0.3\textwidth}
\centering
Input shapes \\ 
\includegraphics[width=0.09\linewidth]{images_supp/arrow.png} \\
Mix parts and generate \\
\end{minipage}
%\hfill
\begin{minipage}{0.65\textwidth}\raggedleft
\centering
\includegraphics[width=\linewidth,trim=2 2 2 2,clip]{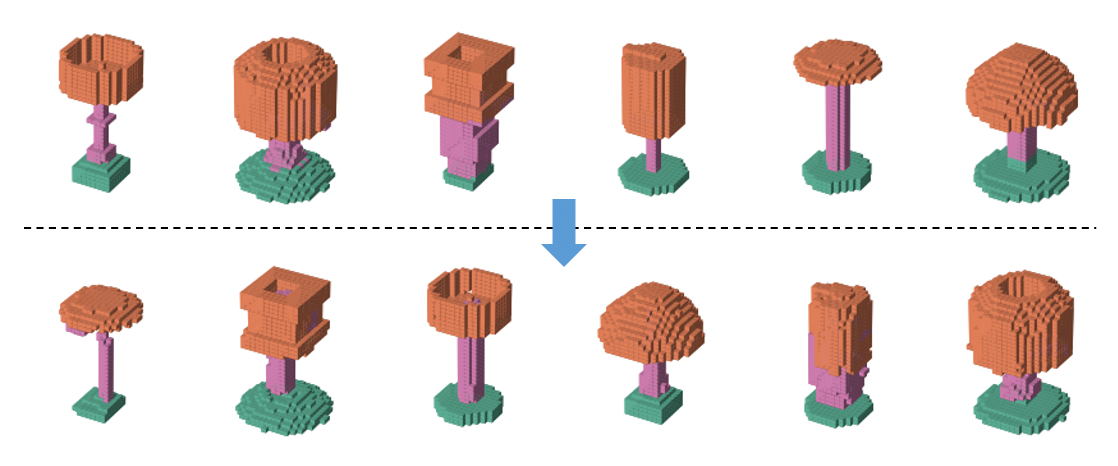}
\end{minipage}
\caption{Part assembly results of mixing random parts from different input shapes of one category to generate new shapes.}
\label{fig:vis_mix}
\end{figure}

Figure \ref{fig:vis_mix} gives some part assembly results of using random parts from the shape category to compose new shapes. From the figure we can observe that for the newly composed shapes, the transformation matrices of different parts are learned coherently for the assembly.

%-------------- Figure: mix -----------------
\begin{figure}[h]
\begin{minipage}{0.3\textwidth}
\centering
Input shapes \\ 
\includegraphics[width=0.09\linewidth]{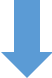} \\
Mix parts and generate \\
\end{minipage}
%\hfill
\begin{minipage}{0.65\textwidth}\raggedleft
\centering
\includegraphics[width=\linewidth,trim=2 2 2 2,clip]{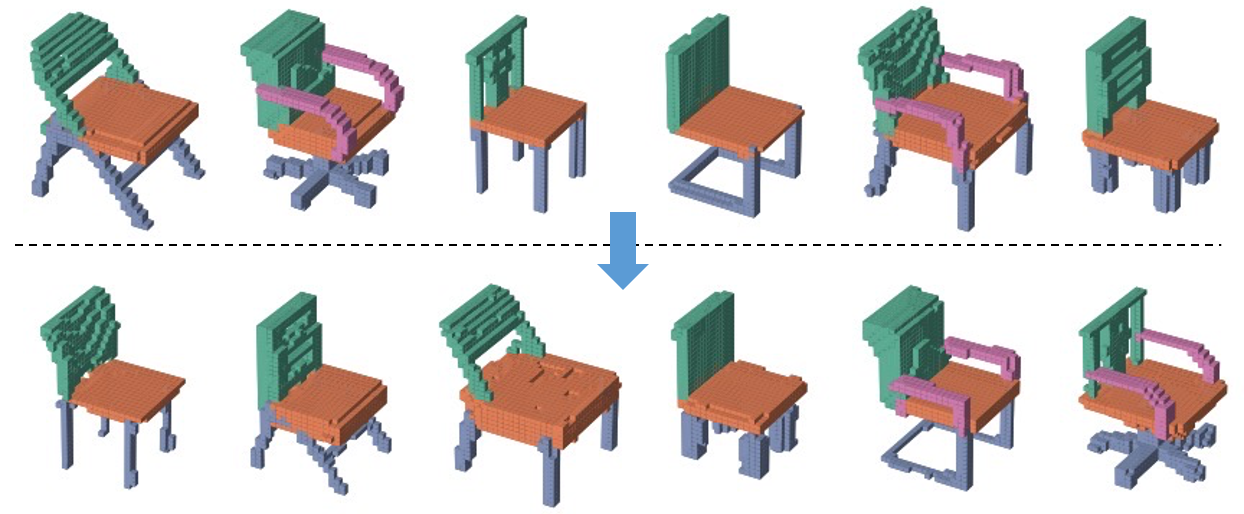}
\end{minipage}
\end{figure}
%\vspace{-1cm}

\begin{figure}[h]
\begin{minipage}{0.3\textwidth}
\centering
Input shapes \\ 
\includegraphics[width=0.09\linewidth]{images_supp/arrow.png} \\
Mix parts and generate \\
\end{minipage}
\begin{minipage}{0.65\textwidth}\raggedleft
\centering
\includegraphics[width=\linewidth,trim=2 2 2 2,clip]{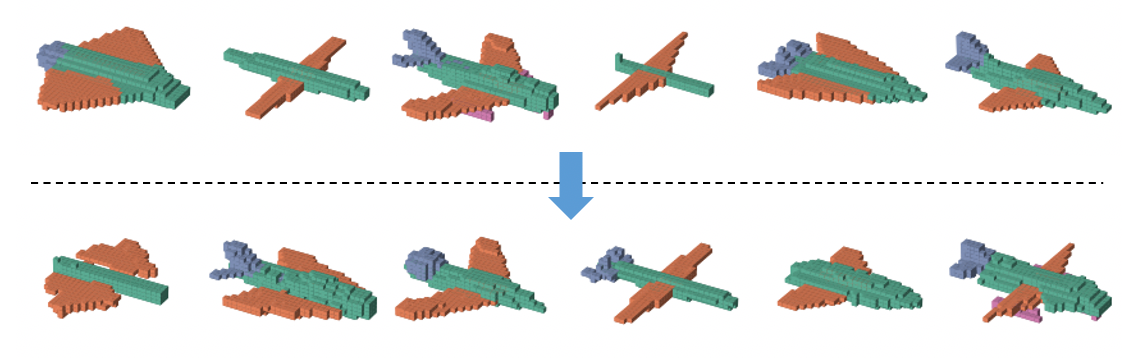}
\end{minipage}
\end{figure}
%\vspace{-1cm}

\begin{figure}[h]
\begin{minipage}{0.3\textwidth}
\centering
Input shapes \\ 
\includegraphics[width=0.09\linewidth]{images_supp/arrow.png} \\
Mix parts and generate \\
\end{minipage}
%\hfill
\begin{minipage}{0.65\textwidth}\raggedleft
\centering
\includegraphics[width=\linewidth,trim=2 2 2 2,clip]{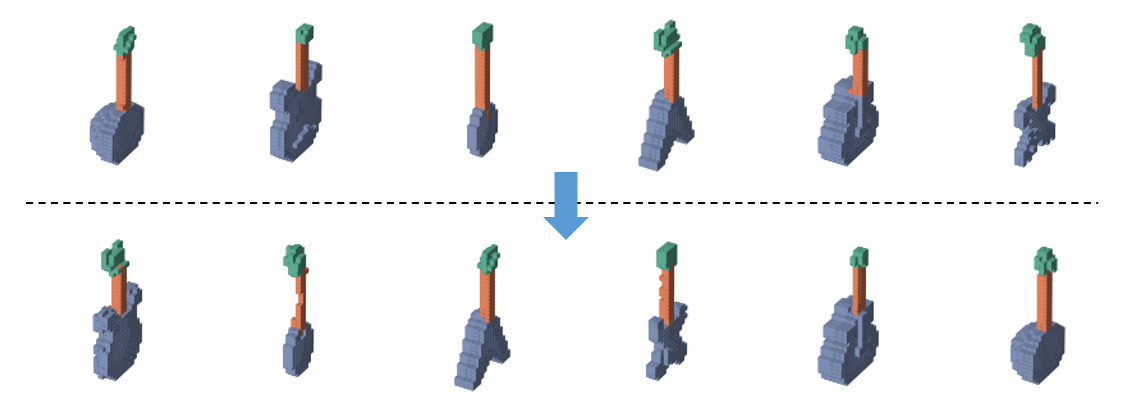}
\end{minipage}
\end{figure}
%\vspace{-1cm}

\clearpage
\subsection{Shape Interpolation}
\begin{figure}[!h]
\centering
\includegraphics[width=0.9\textwidth,trim=2 2 2 2,clip]{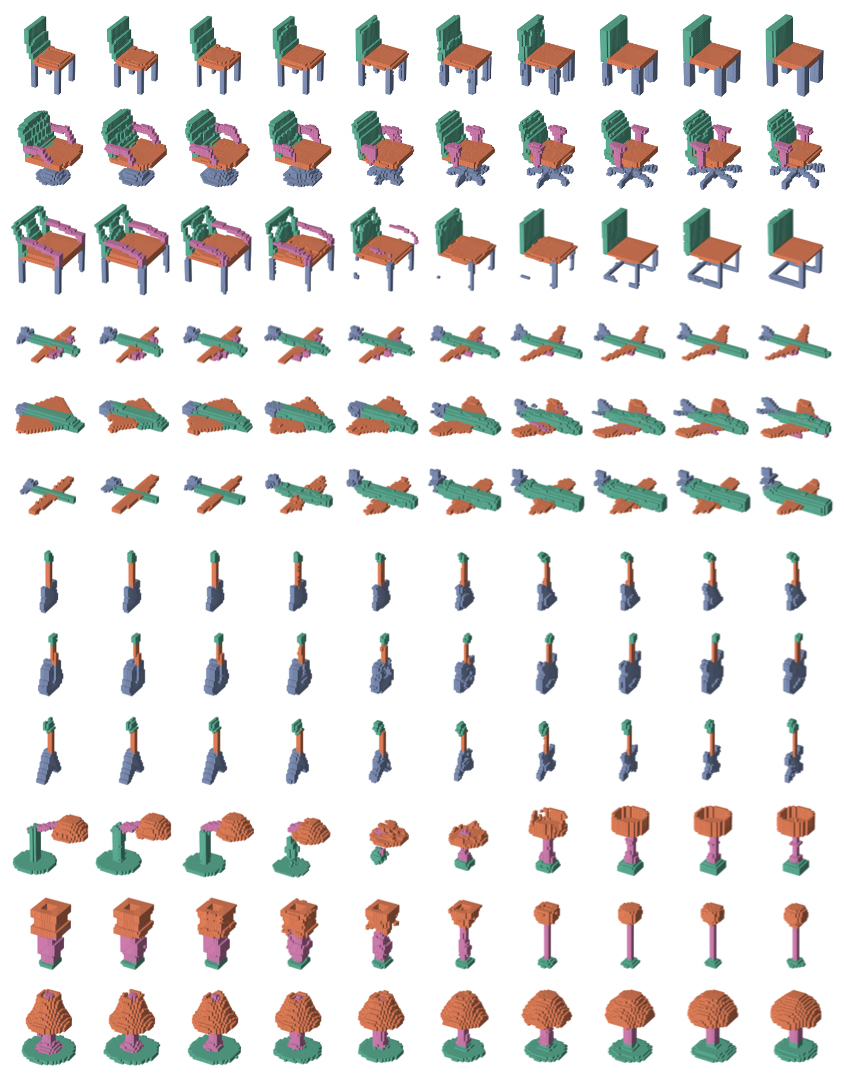}
\caption{Interpolation results between shape pairs from the same category.}
\label{fig:vis_interpolation}
\end{figure}

Additionally, same as most other 3D shape modeling papers, we also give some shape interpolation results as presented in Figure \ref{fig:vis_interpolation}.

\clearpage
\section{Failure Cases}
\begin{figure}[h]
\centering
\includegraphics[width=\textwidth,trim=2 2 2 2,clip]{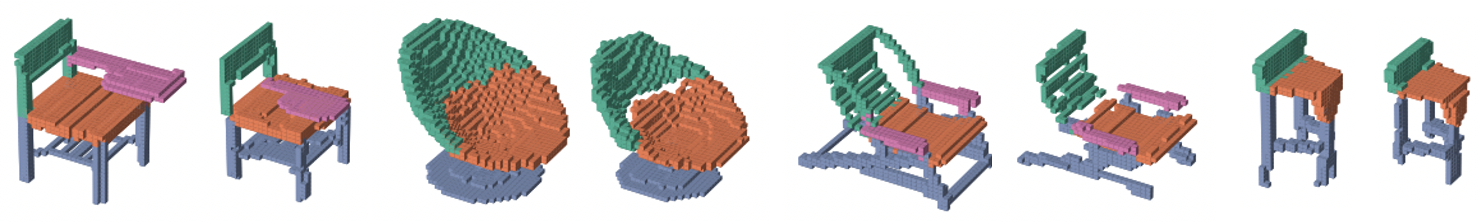}
\caption{Failure cases.}
\label{fig:failureCases}
\end{figure}

We show some failure cases in our experiments in Figure \ref{fig:failureCases}. For each pair, the left shape is the ground truth, while the right shape is the reconstructed one. It can be seen that for some chairs that have uncommon parts, our model fails to reconstruct the parts correctly. However, our model can still assemble the parts into chair-like objects.

\section{Higher Resolution}
\begin{figure}[h]
\centering
\includegraphics[width=0.9\textwidth,trim=2 2 2 2,clip]{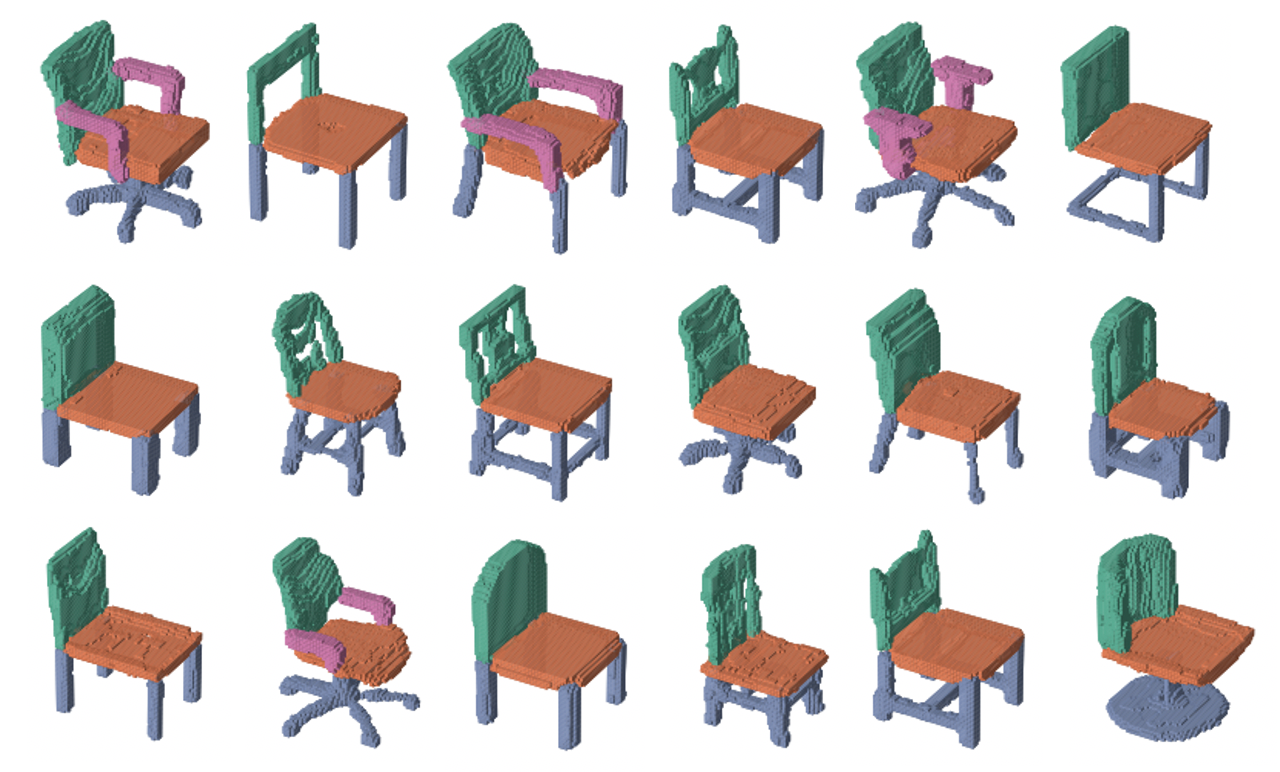}
\caption{Reconstruction results of chair shapes in resolution of $64^3$.}
\label{fig:res64}
\end{figure}

Training a higher resolution of 3D volumetric data could be really time consuming and needs a lot more computation resources. Hence in our paper, all the presented results are from resolution of $32^3$. However, it is surely possible to apply our method to a higher resolution. We hereby show some demo reconstruction results from resolution of $64^3$ in Figure \ref{fig:res64}.

\vfill